\documentclass[lettersize,journal]{IEEEtran}
\usepackage{amsmath,amsfonts}
\usepackage{algorithmic}
\usepackage{algorithm}
\usepackage{array}
\usepackage[caption=false,font=normalsize,labelfont=sf,textfont=sf]{subfig}
\usepackage{textcomp}
\usepackage{stfloats}
\usepackage{url}
\usepackage{verbatim}
\usepackage{graphicx}
\usepackage{cite}
\usepackage{multirow}
\usepackage{colortbl} 
\usepackage[table]{xcolor} 
\usepackage{arydshln}
\usepackage{bbding}

\begin{document}

\title{MM-Retinal V2: Transfer an Elite Knowledge Spark into Fundus Vision-Language  Pretraining}

\author{Ruiqi Wu$^\dag$, 
Na Su$^\dag$, 
Chenran Zhang$^\dag$, 
Tengfei Ma, 
Tao Zhou, % \IEEEmembership{Senior Member, IEEE},
Zhiting Cui, Nianfeng Tang, Tianyu Mao, \\
Yi Zhou$^*$, \IEEEmembership{Senior Member, IEEE}, 
Wen Fan$^*$, 
Tianxing Wu, Shenqi Jing, 
Huazhu Fu,  \IEEEmembership{Senior Member, IEEE}
% \thanks{This work was supported in part by the National Natural Science Foundation of China (Nos. 62476054). Corresponding author: \textit{Yi Zhou} (yizhou.szcn@gmail.com).}
\thanks{R. Wu, C. Zhang, T. Ma, Y. Zhou, and T. Wu are with the School of Computer Science and Engineering, Southeast University, Nanjing, China.}
\thanks{N. Su, Z. Cui, N. Tang, T. Mao, W. Fan, and S. Jing are with the Department of Ophthalmology, The First Affiliated Hospital of Nanjing Medical University, Nanjing, China.}
\thanks{T. Zhou is with the School of Computer Science and Engineering, Nanjing University of Science and Technology, Nanjing, China.}
\thanks{H. Fu is with the Institute of High-Performance Computing, Agency for Science, Technology and Research, Singapore.}}

% The paper headers
\markboth{Journal of \LaTeX\ Class Files,~Vol.~X, No.~X, January~2025}%
{R. Wu \MakeLowercase{\textit{et al.}}: MM-Retinal V2: Transfer an Elite Knowledge Spark into Fundus Vision-Language  Pretraining}

% \IEEEpubid{0000--0000/00\$00.00~\copyright~2021 IEEE}
% Remember, if you use this you must call \IEEEpubidadjcol in the second
% column for its text to clear the IEEEpubid mark.

\maketitle
\def\thefootnote{\dag}\footnotetext{These authors contributed equally to this work.}\def\thefootnote{\arabic{footnote}}
\def\thefootnote{*}\footnotetext{Corresponding authors: \textit{Yi Zhou}(yizhou.szcn@gmail.com) and \textit{Wen Fan}.}\def\thefootnote{\arabic{footnote}}

%%%%%%%%%%%%%%%%%%%%%%%%%%%%%%% Abstract %%%%%%%%%%%%%%%%%%%%%%%%%%%%%%%%%%%%%%%
\begin{abstract}
Vision-language pretraining (VLP) has been investigated to generalize across diverse downstream tasks for fundus image analysis. Although recent methods showcase promising achievements, they significantly rely on large-scale private image-text data but pay less attention to the pretraining manner, which limits their further advancements. In this work, we introduce MM-Retinal V2, a high-quality image-text paired dataset comprising CFP, FFA, and OCT image modalities. Then, we propose a novel fundus vision-language pretraining model, namely KeepFIT V2, which is pretrained by integrating knowledge from the elite data spark into categorical public datasets. Specifically, a preliminary textual pretraining is adopted to equip the text encoder with primarily ophthalmic textual knowledge. Moreover, a hybrid image-text knowledge injection module is designed for knowledge transfer, which is essentially based on a combination of global semantic concepts from contrastive learning and local appearance details from generative learning. Extensive experiments across zero-shot, few-shot, and linear probing settings highlight the generalization and transferability of KeepFIT V2, delivering performance competitive to state-of-the-art fundus VLP models trained on large-scale private image-text datasets. Our dataset and model are publicly available via https://github.com/lxirich/MM-Retinal.
\end{abstract}


%%%%%%%%%%%%%%%%%%%%%%%%%%%%%%% Introduction %%%%%%%%%%%%%%%%%%%%%%%%%%%%%%%%%%%%%%%
\section{Introduction}
\label{sec:introduction}

\IEEEPARstart{F}{undus} imaging serves as a pivotal tool for the examination and diagnosis of ocular diseases. Traditional fundus image analysis models \cite{he2019dme}\cite{wang2020characterization}\cite{peng2021automatic}\cite{zhang2024pyramid} are usually tailored to specific diseases through categorically-labeled training without knowledge integration. These models tend to exhibit limited generalization and transferability. With the considerable advancements in vision-language pretraining, increasing efforts have been put toward the development of fundus foundation models\cite{du2024ret}\cite{mm}\cite{wang2024common}, aiming to build a model capable of diagnosing a wide range of fundus diseases.

To pretrain fundus foundation models, the image-text paired data serve as a critical basis, enabling models to learn aligned vision-language representations that capture numerous ocular diseases' features. However, fundus image-text data are highly scarce. Although a few retinal foundation models\cite{li2024visionunite}\cite{qiu2023visionfm}\cite{wei2024visionclip} have been proposed, the training data they utilized are not released, and most of the existing works only focus on single modality, especially color fundus photography. In real-world clinical diagnosis, different fundus imaging modalities, such as color fundus photography (CFP), fundus fluorescein angiography (FFA), and optical coherence tomography (OCT), are equally important, but collecting large-scale image-text paired data for all the fundus modalities is difficult. Therefore, such limited data acquisition of image-text pairs constrains the development and application of fundus foundation models.

Early fundus foundation models are trained without image-text datasets. RETFound \cite{zhou2023foundation} is trained exclusively on unlabeled images. FLAIR\cite{silva2023foundation} attempts to expand the categorical label of public datasets using templates to generate image-text data. However, both exhibit limited performance. Because of the deficiency of public fundus image-text paired data, recent works\cite{du2024ret}\cite{yang2024vilref}\cite{wang2024common}\cite{shi2024eyeclip} tend to focus on building fundus foundation models by collecting large-scale private image-text pairs. Despite achieving some success, the inherent constraints of these approaches are evident. On the one hand, most of them perform the vision-language pretraining in a brute-force way, and barely emphasize studying the learning manner. On the other hand, the majority of these models are tailored to the CFP modality and trained on private datasets rather than making full use of accumulated public datasets over the past decades, thereby limiting their contribution to the research community.

To address the above problems, we first construct MM-Retinal V2, a high-quality public image-text dataset comprising CFP, FFA, and OCT modalities with around 5K pairs for each modality, and covering over 96 fundus diseases and abnormalities. Enabled by combining MM-Retinal V2 and existing public categorically-labeled datasets, we propose KeepFIT V2, an effective fundus vision-language pretraining method that only requires fewer image-text data resources. The key idea of KeepFIT V2 consists of a preliminary textual knowledge pretraining and a hybrid image-text knowledge injection. In particular, the latter module is performed through a hybrid visual feature matching method which adopts a combination of high-level semantic features derived from contrastive learning and low-level appearance features from generative learning. The two different matching ways complementarily attend the global semantic concepts and local appearance details for knowledge reference from MM-Retinal V2. Finally, an expert knowledge refinement loss is proposed to complete the knowledge injection. Through this training manner, KeepFIT V2 achieves competitive performance with those foundation models pretrained on large-scale private image-text datasets, by leveraging only a small amount of elite image-text data as a knowledge spark. It transfers professional fundus knowledge from MM-Retinal V2 into public datasets that only have categorical labels, and enhances feature alignment and learning during pretraining.

Compared to the previous work MM-Retinal V1 \cite{mm}, in this paper, we make three major aspects of extension. 
\begin{itemize}
\item For the aspect of dataset construction: MM-Retinal V2 significantly expands the CFP and FFA modalities into more than 5K image-text pairs. We also introduce a new OCT modality with a 5K data scale, since the OCT data in \cite{mm} are negligible. Moreover, an MM-Retinal-Text subset essentially from ophthalmology domain is added for text pretraining. 
\item For the aspect of model design: First, compared to KeepFIT V1, preliminary textual knowledge pretraining is employed to enhance the KeepFIT V2 text encoder to better encode retinal knowledge. More importantly, we propose a new hybrid image-text knowledge injection method that combines semantic representation from contrastive learning and appearance representation from generative learning, leading to stronger knowledge transfer. This highlights an effective approach for pretraining a retinal foundation model with limited data resource, delivering performance competitive to models trained on large-scale private image-text datasets.
\item For the aspect of experiment study: Compared to \cite{mm}, more comprehensive and solid experiment evaluations have been conducted, including various experiment settings, compared state-of-the-arts, ablation studies, and more analysis. It shows that the KeepFIT V2 achieves significant improvements. Moreover, experimental results for the new OCT modality are presented in this work.
\end{itemize}

% the features extracted from images are primarily discretized through residual quantization. Then in one route, the discrete visual features are fed into a decoder to reconstruct the image and compute the reconstruction loss; on the other route, we compute the image-text contrastive loss between the discrete visual features and the textual features provided by a text encoder. With this training procedure, the vision tower learns to extract discrete features suitable for both understanding and generation in our VLM.

%%%%%%%%%%%%%%%%%%%%%%%%%%%%%%% Related work %%%%%%%%%%%%%%%%%%%%%%%%%%%%%%%%%%%%%%%
\section{Related Work}
\label{sec:related_work}
\subsection{Retinal Datasets for Disease Diagnosis}
With the growing interest in retinal image diagnosis models, efforts have been made to construct retinal datasets, which can be broadly categorized into two types. (1) Unimodal categorically-labeled datasets\cite{decenciere2014feedback}\cite{porwal2020idrid}\cite{pachade2021retinal} are the most prevalent ones, focusing on disease-specific tasks such as diabetic retinopathy, glaucoma, AMD, and pathologic myopia. These datasets primarily serve as specialized datasets for training disease-specific models. While valuable for image-level classification, these datasets provide limited textual descriptions of the images, restricting their application to foundational pertaining. (2) Image-text paired datasets are crucial for pretraining vision-language models. However, such datasets are scarce due to challenges in acquiring large-scale paired data. Most existing image-text pair datasets\cite{du2024ret}\cite{yang2024vilref}\cite{wang2024common} are private and limited in imaging modality, failing to comprehensively support vision-language research. To address this gap, we publicly release the MM-Retinal V2 dataset, which provides high-quality image-text pairs covering CFP, FFA, and OCT modalities, as detailed in Section~\ref{MM-Retinal_Dataset}.

\subsection{Vision-Language Pre-training}
Vision-language pre-training aims to build relationships between images and texts. There are two mainstream methods. The first category leverages multi-modal encoders based on Transformer to model the interaction between images and texts\cite{chen2023protoclip}\cite{li2023uni}\cite{lu2022unified}\cite{huang2021seeing}. The second category employs an unimodal encoder for images and texts, utilizing contrastive learning to align their representations\cite{radford2021learning}\cite{lavoiemodeling}\cite{li2021align}. In the biomedical domain, PubMedCLIP\cite{eslami2023pubmedclip}, BiomedCLIP\cite{zhang2023biomedclip}, and BioViL\cite{boecking2022making} are proposed as generalist foundation models. Nevertheless, the broad diversity of training data impedes the model's capability to excel in specialized domains like fundus imaging.

\subsection{Retinal Foundational Models}
In addition to generalist biomedical VLP, the rapid advancements in deep learning have brought abundant retinal foundation models to the research area. RETFound\cite{zhou2023foundation} learns from unlabeled retinal images in a self-supervised paradigm. FLAIR\cite{silva2023foundation} utilizes 37 categorical public datasets with textual prompts for foundation model pre-training. However, these works lack image-text paired training data, resulting in limited performance. Recently, RET-CLIP\cite{du2024ret} proposes a binocular pretraining model and ViLRef\cite{yang2024vilref} presents a Chinese vision-language retinal pretraining model, and these two models were trained on over 190K and 450K private clinical retinal images and diagnostic reports, respectively. VisionUnite\cite{li2024visionunite} is tuned on a large private multi-modal fundus dataset which includes over 296K private image-text pairs. RetiZero\cite{wang2024common} develops a private image-text dataset from three sources with ophthalmologists' manual data collection and cleaning. Nonetheless, imaging modalities such as FFA and OCT are also prevalent and of critical importance in real-world clinical practice. All the aforementioned VLP models only support the single CFP modality and the pre-training datasets are not publicly available. Despite some fundus foundation models being designed for multiple image modalities\cite{shi2024eyeclip}\cite{qiu2023visionfm}, they have not been publicly released. Therefore, in this work, in addition to releasing MM-Retinal V2, we also publish pre-trained foundation models for CFP, FFA, and OCT modalities. We aim to provide a method for training VLP models by collaborating small-scale, high-quality image-text paired data with public categorical data in a knowledge-enhanced learning manner by hybrid knowledge injection.

%%%%%%%%%%%%%%%%%%%%%%%%%%%%%%% Dataset %%%%%%%%%%%%%%%%%%%%%%%%%%%%%%%%%%%%%%%
\section{MM-Retinal V2 Dataset}
\label{MM-Retinal_Dataset}

As mentioned in Section~\ref{sec:related_work}, the lack of high-quality image-text paired retinal datasets limits the rapid development of foundation models for fundus imaging. Therefore, we curated a multi-modal dataset comprising retinal image-text pairs in CFP, FFA, and OCT modalities from retinal diagram books and professional assessments provided by ophthalmology experts, named MM-Retinal V2. Meanwhile, to enhance the pre-training of text encoder with comprehensive medical knowledge, particularly in ophthalmology, we constructed a fundus-centric text-only subset named MM-Retinal-Text. 

\subsection{Dataset Construction}
\subsubsection{Image-Text Pairs of CFP and FFA Modalities}
\label{construction_CAP_FFA}

\begin{figure*}[!t]
% \centerline{\includegraphics[width=\columnwidth]{assets/dataset_construction.pdf}}
\centerline{\includegraphics[width=\linewidth]{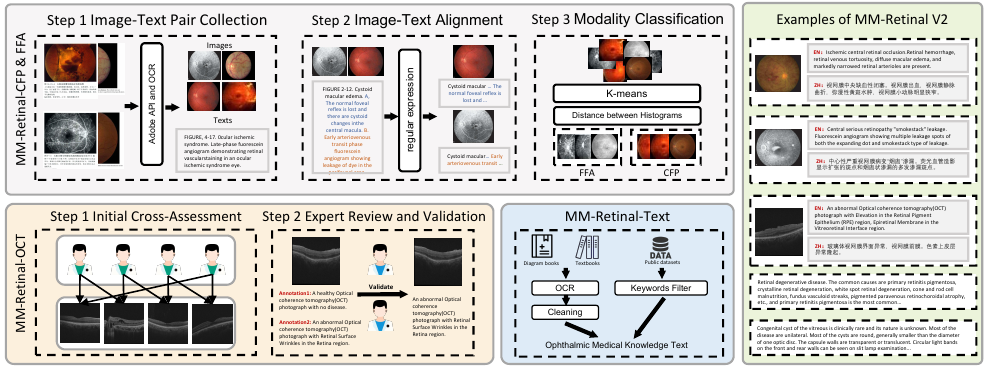}}
\caption{The construction pipeline of MM-Retinal V2 and randomly selected examples from MM-Retinal V2. For CFP and FFA modalities, we propose a semi-automated method consisting of image-text pair collection, image-text alignment, and modality classification. For OCT modality, experienced ophthalmologists are invited for assessment. Each image is annotated twice and finally validated by senior ophthalmologists. To primarily enrich the text encoder with extensive ophthalmic knowledge, we also constructed a text-only subset from fundus diagram books, ophthalmology textbooks, and public datasets for medical LLMs pretraining.}
\label{dataset_construction}
\end{figure*}

All image-text pairs in the CFP and FFA modalities were sourced from four fundus diagram books. As depicted in the upper left of Fig.~\ref{dataset_construction}, a semi-automated three-stage pipeline was implemented to construct the dataset. First, we manually captured all image-text pairs from the diagram books, ensuring each pair was recorded in a single screenshot with a resolution of no less than 800×800 pixels for each image. These screenshots were then processed using a custom program that integrates Adobe tools for image extraction and OCR technology for text extraction. 
Then, given the challenge of aligning extracted images and texts—especially in cases where multiple sub-figures are associated with a single caption—we applied regular expressions to achieve precise caption separation. 
Finally, the images were classified into CFP and FFA modalities using a color histogram-based method. We manually corrected OCR recognition errors and translated the text into English and Chinese reports to ensure language consistency. 
To further expand our dataset, we refined the DEN dataset\cite{huang2021deepopht} through a systematic filtering and cleaning process. This involved manually removing images that did not belong to the CFP, and FFA modalities, as well as excluding collages comprising multiple images. Subsequently, a color histogram-based classification method was used to automatically classify two modalities, and the label files were reorganized according to their respective modalities.

\subsubsection{Image-Text Pairs of OCT Modality}
Due to the scarcity of electronic OCT diagram books with rich image-text paired data, we collaborated with a key provincial hospital to establish the OCT modality part of MM-Retinal V2. % The data collection took place in Nanjing, China, between xx/xxxx and xx/xxxx. Approval was obtained from the Institutional Review Board (xxxx) and the Ethics Committee of xxxx. 
We initially collected 5,587 OCT images from 3,403 patients. To control data quality, images that could not be diagnosed solely based on OCT were excluded. The ophthalmologist provided detailed definitions for the full range of abnormalities that can be observed from OCT images. Then, each image was independently assessed by two ophthalmologists. Two senior ophthalmologists reviewed all captions and provided final verification and decisions, ensuring the correctness and consistency of the dataset. At last, each OCT image was accompanied by a detailed textual description, capturing both the diagnosed diseases and the pathological features observed in the images. 

\subsubsection{Ophthalmic Knowledge Texts}
\label{medical_text}

When diagnosing ocular diseases, ophthalmologists usually rely not only on ophthalmic images but also on their knowledge of ophthalmology and other medical specialties. To support this, we also constructed an MM-Retinal-Text subset, which primarily integrates knowledge from the ophthalmic field. 
The textual data were sourced from three main origins: (1) four fundus diagram books, (2) three ophthalmology textbooks, and (3) twelve public datasets for medical LLM pretraining. Texts from books were digitized using OCR, with irrelevant elements such as names and figure sequence numbers removed. For public datasets used in medical LLM pretraining, we filtered them using ophthalmology-related keywords derived from the contents of diagram books, as these datasets were originally designed for comprehensive medical areas. Although this filtering approach ensures a focus on ophthalmic knowledge, it may also include content from other medical specialties. We opted not to clean this data further, as a certain proportion of knowledge from other fields would improve the model’s capability to generalize on common medical terms. The detailed experimental results are presented in Table~\ref{ablation_study_table}. More details on these datasets can be found in our project page.

\subsection{Dataset Statistics}

\begin{figure}[!t]
\centerline{\includegraphics[width=\columnwidth]{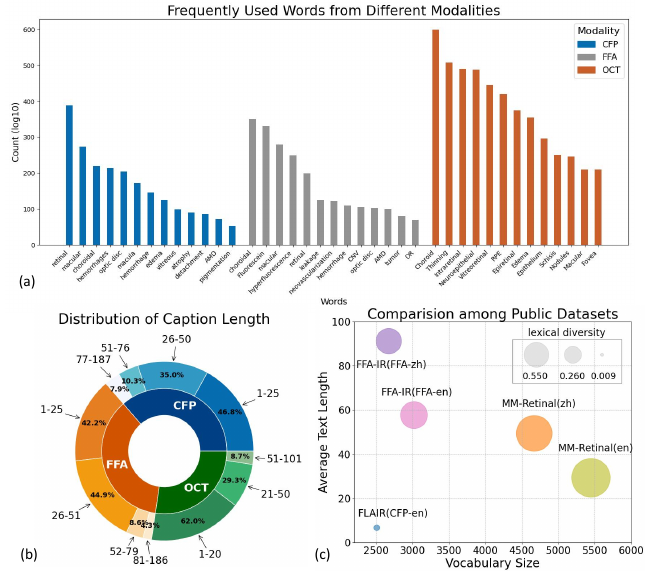}}
\caption{Statistical overview of MM-Retinal V2. (a) highlights a part of the most frequently occurring terms in the CFP, FFA, and OCT modalities. (b) plots the distribution of caption length in three modalities, respectively. (c) illustrates the comparison with related public fundus datasets in the aspect of vocabulary. The average vocabulary size refers to the total number of unique words throughout all captions in the dataset, while lexical diversity measures the average number of unique words used in each individual caption.}
\label{dataset_statistics}
\end{figure}

Upon MM-Retinal V1 \cite{mm}, our MM-Retinal V2 finally consists of 6,720 CFP cases, 5,119 FFA cases, and 5,502 OCT cases, each containing an image paired with corresponding descriptions in both Chinese and English. The MM-Retinal V2 dataset is further enriched by a text-only subset, with a total number of 452K utterances, providing a substantial repository of textual information. 

\subsubsection{Diverse Modalities}
MM-Retinal V2 is the first high-quality dataset to simultaneously encompass image-text pairs of CFP, FFA, and OCT modalities with ophthalmic text data. In the diagnostic process, different imaging modalities offer unique perspectives. CFP highlights the structure of the fundus, FFA captures vascular changes, while OCT reveals details of the retinal layers. By integrating these modalities with high-quality textual descriptions, MM-Retinal V2 provides a solid base for the development of advanced retinal foundational models.

\subsubsection{Comprehensive Categories} 
As the MM-Retinal V2 dataset is derived from comprehensive ocular diagram books and clinical sources, it comprises over 96 fundus abnormalities and disease categories, including both common and rare diseases, such as retinal vascular diseases, macular diseases, vitreous diseases, optic nerve diseases, congenital anomalies, and inflammatory diseases, among others. Fig.~\ref{dataset_statistics} (a) shows the words that appear most frequently. More detailed retinal categories are presented in the supplementary material, providing a broader perspective on the extensive coverage of the dataset.

\subsubsection{Detailed Captions}
Fig.~\ref{dataset_statistics} (b) shows the distribution of caption lengths in MM-Retinal V2. In the CFP modality, 46.8\% and 35.0\% of captions range from 1 to 25 words and 26 to 50 words. In the FFA modality, 57.8\% caption length is over 25 words. In the OCT modality, 91.3\% captions range from 1 to 50 words. Moreover, a small percentage of texts exceed 50 words, reaching up to nearly 101 words. This demonstrates that the textual captions in MM-Retinal V2 are detailed and comprehensive, providing accurate and rich descriptions of the images and effectively conveying the information appearing.

\subsubsection{Extensive Vocabulary}
MM-Retinal V2 captions encompass disease diagnoses, detailed lesion attributes (such as color, shape, and appearance), clinical symptoms, and post-treatment efficacy, utilizing a rich vocabulary for comprehensive descriptions. Fig.~\ref{dataset_statistics}(c) compares the vocabulary size and lexical diversity across various public datasets and languages. Notably, since MM-Retinal V2 does not generate captions by merely expanding category names with fixed templates like \cite{silva2023foundation}, its vocabulary is exceptionally rich and diverse, exhibiting high lexical diversity.

%%%%%%%%%%%%%%%%%%%%%%%%%%%%%%% Method %%%%%%%%%%%%%%%%%%%%%%%%%%%%%%%%%%%%%%%
\section{KeepFIT V2}

\begin{figure*}[!t]
% \centerline{\includegraphics[width=\columnwidth]{assets/dataset_construction.pdf}}
\centerline{\includegraphics[scale=1.3]{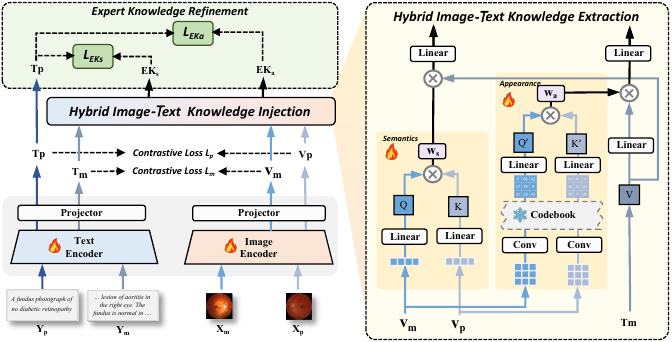}}
\caption{The architecture of the proposed KeepFIT V2. KeepFIT V2 generally complies with the vision-language pretraining paradigm and introduces four specific parts, including a preliminary textual knowledge pretraining, a semantics-oriented knowledge extraction module, an appearance-oriented knowledge extraction module, and an expert knowledge refinement module. The image encoder and the text encoder extract features and encourage modality alignment by contrastive learning. The semantics-oriented and appearance-oriented knowledge extraction modules are introduced to distill expert knowledge from MM-Retinal V2. Subsequently, the text refinement module injects the obtained knowledge into public datasets to enhance model pretraining.}
\label{model_framework}
\end{figure*}

In this section, we introduce the proposed KeepFIT V2, a new knowledge-enhanced multi-modal foundation model designed for retinal image analysis. Compared to the previous version \cite{mm}, KeepFIT V2 firmly follows the vision-language pretraining paradigm and adopts a more effective hybrid image-text knowledge injection approach by leveraging the high-quality MM-Retinal V2 dataset. Therefore, such an elite knowledge spark can be transferred into the general vision-language pertaining to enhance model performance. The framework of KeepFIT V2 is illustrated in Fig.~\ref{model_framework}.
 
\subsection{Vision-Language Pretraining Framework}
KeepFIT V2 is trained on MM-Retinal V2 $m$ and public retinal datasets $p$ that only encompass category-level labels. Due to the effectiveness of vision-language pretraining paradigm, CLIP\cite{radford2021learning} is applied as the backbone of KeepFIT V2 for multi-modal learning. Specifically, KeepFIT V2 comprises two encoders. Provided with a set of image-text pairs $\{X_i, Y_i\}_{i=1}^N$, where $X_i$ represents the image, and $Y_i$ is the corresponding text, the image is processed by the vision encoder $E_v$ to extract the visual feature $V_i$, while the text is fed into the text encoder $E_t$ to obtain the textual feature $T_i$. For categorical public datasets, a template is needed to convert the category label into text, such as ``A fundus photograph of [class name]'' for CFP modality, and a label augmentation is applied following FLAIR\cite{silva2023foundation}. Then, the features are projected to a shared space by modality-specific projector $P_v$ and $P_t$ to ensure that the feature dimensions $d$ of different modalities are consistent for contrastive learning. Let $\theta$ and $\phi$ symbolize the parameters of the image encoder and text encoder, respectively. The generated image feature $V_i$ and text feature $T_i$ can be formulated as follows:
\begin{equation}
V_i = P_{v} \circ E_{v}(X_i;\theta) \in \mathbb{R^{d}}^{d},~~ T_i = P_{t} \circ E_{t}(Y_i;\phi) \in \mathbb{R^{d}}^{d}.
\label{eq_clip}
\end{equation}

To eliminate the modality gap within the shared space after projection, we use contrastive loss to enable the modality alignment ability. Give an image-text pair, image-to-text and text-to-image similarities after applying a softmax function are calculated by: 
\begin{align}
U_{v2t}(V_i)=\frac{exp(S(V_i,T_i)/\tau)}{\sum_{j=1}^{\mathcal{B}}exp(S(V_i,T_j)/\tau)},\\ 
U_{t2v}(T_i)=\frac{exp(S(T_i,V_i)/\tau)}{\sum_{j=1}^{\mathcal{B}}exp(S(T_i,V_j)/\tau)},
\label{eq_similarity}
\end{align}
where $S(\cdot,\cdot)$ refers to cross-modality similarity, $\tau$ is a temperature parameter, and $\mathcal{B}$ is batch size. Then, the image-text contrastive loss is defined as:
\begin{equation}
\label{contrastive_loss}
  \begin{aligned}
    \mathcal{L}_{itc} & =\frac{1}{2}\mathbb{E}_{(V,T)\sim\mathcal{B}}[CE(G_{v2t}(V),U_{v2t}(V)) \\
    & + CE(G_{t2v}(T),U_{t2v}(T))].
  \end{aligned}
\end{equation}

To achieve better image-text alignment, Eq.(4) is employed on MM-Retinal V2 and categorical public datasets. For MM-retinal V2, the matching labels $G$ are identity matrices of dimension $\mid \mathcal{B} \mid \times \mid \mathcal{B} \mid$. On the other hand, for public datasets that only provide category labels, negative samples for the contrastive task come from different categories. Thus, the matching labels are symmetric matrices of size $\mid \mathcal{B} \mid \times \mid \mathcal{B} \mid$.

\subsection{Preliminary Textual Knowledge Pretraining}
\label{textual_knowledge_pretraining}
Following FLAIR\cite{silva2023foundation}, we adopt ResNet50\cite{he2016deep} initialized with ImageNet pre-trained parameters as the image encoder, and choose the architecture of BioClinicalBert\cite{alsentzer2019publicly} as the text encoder. As mentioned in Section~\ref{medical_text}, extensive professional texts in the medical field, especially ophthalmology, serve as a preliminary expert knowledge source for the diagnosis of fundus diseases. Hence, we propose to exploit such abundant and profound knowledge from MM-Retinal-Text to better adapt the text encoder of KeepFIT V2. Specifically, we pretrain the text encoder using MM-Retinal-Text through the masked language modeling (MLM) fashion. The pre-trained parameters are used to initialize the text encoder. Experiments in Table~\ref{ablation_study_table} show that this training step yields stronger performance and enhances the text encoder’s capability to capture intricate medical terminology and context, thus facilitating better alignment with the image encoder in our KeepFIT V2 framework.

\subsection{Hybrid Image-Text Knowledge Injection}
In light of the MM-Retinal V2 dataset incorporating a wealth of fundus image-text expert knowledge, we consider injecting expert knowledge from MM-Retinal V2 into public datasets to promote vision-language pretraining. By jointly training the model on MM-Retinal V2 and public datasets, we seek to enhance the model's understanding of fundus images and related textual information.

To accomplish this, three key obstacles must be addressed. The first question is what the expected transferable knowledge for extraction should be and where the extracted knowledge is injected. The second focuses on how to extract knowledge effectively. The third is how to inject the extracted knowledge. In the following subsections, we will systematically address these problems.

\subsubsection{Elite Knowledge Spark}
By comparing the images from two data sources, namely public datasets and our MM-Retinal V2 dataset, we observe that MM-Retinal V2 exhibits a high degree of similarity with the public ones and almost covers all the common retinal disease categories. However, the textual content of the public datasets contains simple expansions of category labels based on fixed text templates, while the texts in MM-Retinal V2 are extensive and lexically diverse. Textual discrepancies reveal where knowledge dissemination is most needed. Consequently, to address the first obstacle, the image-guided texts of the MM-Retinal V2 will be regarded as the source for knowledge extraction, and the textual content of public datasets as the destination where knowledge is injected. This allows the knowledge from the MM-Retinal V2 to act as a spark, spreading and enriching the public datasets, and jointly contribute to the pretraining of KeepFIT V2.

\subsubsection{Semantics-Oriented Expert Knowledge Extraction}
We next investigate how to extract the expert knowledge from MM-Retinal V2. As mentioned above, MM-Retinal V2 shows minimal domain differences with public datasets and encompasses their fundus disease categories, making it ideal for knowledge extraction. Hence, we propose a semantics-oriented expert knowledge extraction module based on multi-head cross attention\cite{vaswani2017attention} to equip the model with high-level semantic retrieval ability. In particular, we perform visual matching on the images from two data sources and weight the corresponding text from the MM-Retinal V2 as expert knowledge based on the matching score.

Given the image feature $V_p$, $V_m$ from image encoder $E_v$ and text feature $T_p$, $T_m$ from text encoder $E_t$, we input the image features of public datasets as $query$ of cross attention, the image features of the MM-Retinal V2 as $key$, and the text features of the MM-Retinal V2 as $value$. Subsequently, matrix multiplication followed by the softmax function is applied to compute the attention weights $\Psi^{h}$ for head $h\in[H]$, which represent similarity scores between images:
\begin{align}
& Q_{i}^{h}=V_{p,i} \cdot W_{Q}^{h} \in \mathbb{R}^{d/H}, \\
& K_{j}^{h}=V_{m,j} \cdot W_{K}^{h} \in \mathbb{R}^{d/H}, \\
& \Psi^{h}_{ij}=softmax(\frac{Q_{i}^{h}{K_{j}^{h}}^T}{\sqrt{d/H}} ),
\label{attention_weight}
\end{align}
where $p$ symbolizes public datasets that only have category labels and $m$ indicates MM-Retinal V2 dataset, and $W_Q^h$, $W_K^h$ are learnable projection matrices.

Next, similarity scores are used to reweight the text features of MM-Retinal V2, assigning different levels of attention to the text features based on the semantic similarity between image features. The semantics-level expert knowledge $EK_s$ from MM-Retinal V2 can be formulated as:
\begin{align}
& V_{j}^{h}=T_{m,j} \cdot W_{V}^{h} \in \mathbb{R}^{d/h}, \\
& EK_s^h=\Psi^{h}_{ij} \cdot V^h_{j}, \\
& EK_s=concat(EK_s^h)_{h=1}^H \cdot W_O,
\label{extracted_knowledge}
\end{align}
where $W_V^h$ and $W_O$ are learnable projection matrices.

When images from the public datasets and the MM-Retinal V2 demonstrate high similarity, their corresponding texts are expected to align more closely. As a result, higher similarity scores lead to greater text retention from MM-Retinal V2, thereby enriching the public datasets with relevant and complementary textual knowledge in a semantic visual matching way.

\subsubsection{Appearance-Oriented Expert Knowledge Extraction}
In addition to high-level semantic visual matching, we further explore matching the visual features between public datasets and MM-Retinal V2 in a low-level appearance way. Detailed appearance features can be derived from image generative learning. Therefore, a combination of semantic representation from contrastive learning and appearance representation from generative learning is a complementary way to achieve this goal. Specifically, we employ a vector quantization (VQ) approach to converting semantic visual features into discrete tokens for appearance-oriented knowledge extraction. 

Fully representing continuous retinal features from the image encoder of KeepFIT V2 by quantized discrete tokens is challenging, as these retinal image features are highly semantic and contain limited low-level visual information like detailed lesion appearance. To address this, image tokenization in appearance-oriented knowledge extraction is inspired by index backpropagation quantization (IBQ) \cite{shi2024taming}, which leverages a large-scale, high-dimensional codebook with efficient utilization. Moreover, IBQ allows the joint optimization of all codebook embeddings, effectively preventing codebook collapse. In this work, the codebook is trained in advance, following \cite{shi2024taming}, using the same training data as KeepFIT V2, which includes images from MM-Retinal V2 and public retinal datasets.

As illustrated in the lower right of Fig.~\ref{model_framework}, after obtaining the image features $V$ from the image encoder, these image features comprise two parts, with the first part being the flattened features processed through the projector $P_v$ and the second part being unflattened feature maps bypassing the projector.
The visual feature maps are first projected by a convolutional layer to achieve dimensional consistency. Then a quantization process is performed to tokenize the contiguous visual feature maps into discrete tokens using a fixed codebook $C \in \mathbb{R}^{K \times D}$, where $K$ is the codebook size and $D$ is the code dimension. First, the dot product between the visual feature and all code embeddings $C_k$ is calculated and followed by the softmax function and one-hot function to obtain probabilities:
\begin{align}
& logits=[V^TC_1,V^TC_2,...,V^TC_K]^T\in \mathbb{R}^K, \\
& Ind_{soft}=softmax(logits), \\
& Ind_{hard}=OneHot(argmax(Ind_{soft})).
\label{IBQ_1}
\end{align}

Afterward, the gradients of soft one-hot distribution are transferred to hard one-hot index:
\begin{equation}
Ind=Ind_{hard}-sg[Ind_{soft}]+Ind_{soft},
\label{IBQ_2}
\end{equation}
where $sg[\cdot]$ means stop-gradient operation.

After acquiring the index, the discrete code $\mathcal{Q}$ obtained by IBQ is:
\begin{equation}
\mathcal{Q}=Ind^TC.
\label{IBQ_3}
\end{equation}

Similarly, multi-head cross-attention is leveraged for knowledge extraction in an appearance-oriented way. The quantized vector $\mathcal{Q}_p$ from the public datasets is used as the $query$, $\mathcal{Q}_m$ from the MM-Retinal V2 as the $key$, and the text feature $T_m$ from the MM-Retinal V2 as the $value$. Consequently, Eqs. (5) and (6) are changed into:
\begin{align}
& Q_{i}^{h}=\mathcal{Q}_{p,i} \cdot W_{Q}^{h} \in \mathbb{R}^{d'/H}, \\
& K_{j}^{h}=\mathcal{Q}_{m,j} \cdot W_{K}^{h} \in \mathbb{R}^{d'/H}.
\end{align}

The final extracted appearance-level expert knowledge $EK_a$ is computed by:
\begin{align}
& EK_a^h=\sigma_{\tau}(Q^{h}_{i} \cdot K^{h}_{j}) \cdot V^h_j, \\
& EK_a=concat(EK_a^h)_{h=1}^H \cdot W_O,
\end{align}
where $\sigma_{\tau}$ denotes the softmax function with temperature $\tau$.

The hybrid expert knowledge extraction module empowers KeepFIT V2 with both global semantic representation ability from contrastive learning and local appearance representation ability from generative learning, which is one of the major improvements compared to KeepFIT V1. Through this module, the cross-modality alignment capability and representation capability of KeepFIT V2 are significantly enhanced, resulting in more precise visual matching and knowledge injection.

\subsubsection{Expert Knowledge Refinement}
The last significant problem is how to inject the obtained knowledge into the text of the public datasets to assist vision-language pretraining. Considering that the primary distinction between the public datasets and MM-Retinal V2 lies in the depth and granularity of their texts, we propose expert knowledge refinement loss $L_{EK_s}$ for semantics-oriented knowledge refinement and $L_{EK_a}$ for appearance-oriented knowledge refinement. These losses encourage the text of the public datasets to closely resemble the text extracted from MM-Retinal V2 that corresponds to their image features. To achieve this, the mean squared error (MSE) Loss is utilized as the basis:
\begin{align}
& \mathcal{L}_{EK}^{s}=\frac{1}{\mathcal{B}}\sum_{i=1}^{\mathcal{B}}(EK_s-T_p)^2, \\
& \mathcal{L}_{EK}^{a}=\frac{1}{\mathcal{B}}\sum_{i=1}^{\mathcal{B}}(EK_a-T_p)^2.
\label{LEKs}
\end{align}

The above formulas use the knowledge extracted at two levels to refine the text of the public datasets, creating a complementary and synergistic effect that makes the text refinement more comprehensive.

\begin{table*}[]
\caption{Zero-Shot Classification Performance in CFP Modality across Four datasets. (\%)}
\label{CFP_zeroshot}
\setlength{\tabcolsep}{5.6pt}
\renewcommand\arraystretch{1.2}
\begin{tabular}{ccccccccccccccccc}
\hline
\multirow{2}{*}{Model} & \multicolumn{4}{c|}{REFUGE}                        & \multicolumn{4}{c|}{ODIR200$\times$3}                          & \multicolumn{4}{c|}{Retina}                        & \multicolumn{4}{c}{iChallenge-AMD}       \\ \cline{2-17} 
                       & ACC   & AUC   & AUPR  & \multicolumn{1}{c|}{AVG}   & ACC   & AUC   & AUPR  & \multicolumn{1}{c|}{AVG}   & ACC   & AUC   & AUPR  & \multicolumn{1}{c|}{AVG}   & ACC   & AUC   & AUPR  & AVG   \\ \hline
\multicolumn{17}{l}{\textit{VLPs with large-scale image-text paired data}}                            \\ \hdashline
ViLRef                 & 64.3          & 76.7          & 69.7          & \multicolumn{1}{c|}{70.2}          & 88.3          & 96.2          & 92.9          & \multicolumn{1}{c|}{92.5}          & 54.6          & 80.7          & 62.6          & \multicolumn{1}{c|}{66.0}          & 84.3          & 95.2          & 94.0          & 91.2          \\
RET-CLIP                                                   & 81.0          & 94.6          & 90.0          & \multicolumn{1}{c|}{88.5}          & 88.2          & 96.6          & 93.0          & \multicolumn{1}{c|}{92.6}          & 61.7          & 89.6          & 76.6          & \multicolumn{1}{c|}{76.0}          & 87.2          & 94.5          & 93.2          & 91.6          \\
RetiZero                                                   & 53.3          & 82.5          & 70.5          & \multicolumn{1}{c|}{68.8}          & 71.3          & 97.9          & 96.0          & \multicolumn{1}{c|}{88.4}          & 42.4          & 78.6          & 58.6          & \multicolumn{1}{c|}{65.0}          & 66.6          & 87.9          & 87.0          & 80.5 \\ \hline
\multicolumn{17}{l}{\textit{VLPs with small-scale elite image-text paired data / public categorical data}}                                        \\ \hdashline
FLAIR                  & 84.7          & 92.6          & 90.5          & \multicolumn{1}{c|}{89.3}          & 40.3          & 87.5          & 76.9          & \multicolumn{1}{c|}{68.2}          & 33.8          & 69.9          & 45.9          & \multicolumn{1}{c|}{49.9}         & 69.5          & 79.5          & 75.7          & 74.9          \\
KeepFITV1                                                  & 84.9          & 94.1          & 89.3          & \multicolumn{1}{c|}{89.4}          & 81.2          & 92.9          & 87.5          & \multicolumn{1}{c|}{87.2}          & 42.9          & 77.4          & 52.0          & \multicolumn{1}{c|}{57.4}          & 76.5          & 88.6          & 86.2          & 83.8          \\
\rowcolor[HTML]{EFEFEF} % 设置浅灰色背景
KeepFITV2              & \textbf{89.6} & 96.2          & 92.7          & \multicolumn{1}{c|}{\textbf{92.8}} & 80.8          & 93.1          & 87.6          & \multicolumn{1}{c|}{87.2}          & 43.6          & \textbf{80.8} & \textbf{58.8} & \multicolumn{1}{c|}{\textbf{61.1}} & 80.4          & 90.3          & 87.1          & 85.9          \\
\rowcolor[HTML]{DCDCDC} % 设置浅灰色背景
KeepFITV2$_L$              & 86.2          & \textbf{96.9} & \textbf{94.8} & \multicolumn{1}{c|}{92.6}          & \textbf{85.2} & \textbf{94.8} & \textbf{90.3} & \multicolumn{1}{c|}{\textbf{90.1}} & \textbf{45.1} & 78.1          & 57.9          & \multicolumn{1}{c|}{60.4}          & \textbf{82.7} & \textbf{91.2} & \textbf{90.2} & \textbf{88.0}  \\ \hline
\end{tabular}
\end{table*}

\subsubsection{Overall Training Objective}
As depicted in the Fig.~\ref{model_framework}, the pretrained codebook is frozen, and we optimize the parameters of the image encoder, text encoder, and hybrid knowledge extraction modules, simultaneously. Finally, the overall training objective is defined as:
\begin{equation}
\mathcal{L}=\mathcal{L}_{itc}^{p} + \mathcal{L}_{itc}^{m} + \lambda_1\mathcal{L}_{EK}^{s} + \lambda_2\mathcal{L}_{EK}^{a},
\label{training_objective}
\end{equation}
where $\lambda_1$ and $\lambda_2$ are hyperparameters, which are set to 100 and $1\times10^4$ in our implementation, achieving the best performance.

%%%%%%%%%%%%%%%%%%%%%%%%%%%%%%% Experiments %%%%%%%%%%%%%%%%%%%%%%%%%%%%%%%%%%%%%%%
\section{Experiments}
\label{sec:experiments}

\begin{table*}[]
\caption{Few-Shot Classification Performance in CFP Modality across Four datasets. (\%)}
\label{CFP_fewshot}
\setlength{\tabcolsep}{0.5pt}
\renewcommand\arraystretch{1.2}
\begin{tabular}{ccccccccccccccccccccccccccccc}
\hline
\multirow{3}{*}{Model} & \multicolumn{9}{c|}{Clipadapter}                                                        & \multicolumn{9}{c|}{Tipadapter}                                                         & \multicolumn{9}{c|}{Tipadapter-f}                                                       & \multirow{3}{*}{AVG} \\
                       & \multicolumn{3}{c}{1} & \multicolumn{3}{c}{5} & \multicolumn{3}{c|}{10}                 & \multicolumn{3}{c}{1} & \multicolumn{3}{c}{5} & \multicolumn{3}{c|}{10}                 & \multicolumn{3}{c}{1} & \multicolumn{3}{c}{5} & \multicolumn{3}{c|}{10}                 &                      \\ \cline{2-28}
                      & {\fontsize{7pt}{12pt}\selectfont ACC}   & {\fontsize{7pt}{12pt}\selectfont AUC}   & {\fontsize{7pt}{12pt}\selectfont AUPR}  & {\fontsize{7pt}{12pt}\selectfont ACC}   & {\fontsize{7pt}{12pt}\selectfont AUC}   & {\fontsize{7pt}{12pt}\selectfont AUPR}  & {\fontsize{7pt}{12pt}\selectfont ACC}   & {\fontsize{7pt}{12pt}\selectfont AUC}   & \multicolumn{1}{c|}{{\fontsize{7pt}{12pt}\selectfont AUPR}}  & {\fontsize{7pt}{12pt}\selectfont ACC}   & {\fontsize{7pt}{12pt}\selectfont AUC}   & {\fontsize{7pt}{12pt}\selectfont AUPR}  & {\fontsize{7pt}{12pt}\selectfont ACC}   & {\fontsize{7pt}{12pt}\selectfont AUC}   & {\fontsize{7pt}{12pt}\selectfont AUPR}  & {\fontsize{7pt}{12pt}\selectfont ACC}   & {\fontsize{7pt}{12pt}\selectfont AUC}   & \multicolumn{1}{c|}{{\fontsize{7pt}{12pt}\selectfont AUPR}}  & {\fontsize{7pt}{12pt}\selectfont ACC}   & {\fontsize{7pt}{12pt}\selectfont AUC}   & {\fontsize{7pt}{12pt}\selectfont AUPR}  & {\fontsize{7pt}{12pt}\selectfont ACC}   & {\fontsize{7pt}{12pt}\selectfont AUC}   & {\fontsize{7pt}{12pt}\selectfont AUPR}  & {\fontsize{7pt}{12pt}\selectfont ACC}   & {\fontsize{7pt}{12pt}\selectfont AUC}   & \multicolumn{1}{c|}{{\fontsize{7pt}{12pt}\selectfont AUPR}}  &                      \\ \hline
\multicolumn{29}{c}{REFUGE} \\ \hline
\multicolumn{29}{l}{\textit{VLPs with large-scale image-text paired data}}    \\ \hdashline  
ViLRef                  & 78.1          & 88.1          & 83.6          & 81.2          & 89.0          & 87.5          & 83.5          & 90.1          & \multicolumn{1}{c|}{88.0}          & 62.9          & 69.5          & 64.1          & 66.0          & 72.2          & 65.3          & 69.3          & 75.2          & \multicolumn{1}{c|}{67.6}          & 60.3          & 67.7          & 62.8          & 68.4          & 76.4          & 65.5          & 77.9          & 82.2          & \multicolumn{1}{c|}{73.4}          & 70.5                 \\
RET-CLIP                                                  & 84.0          & 91.5          & 88.2          & 80.7          & 90.2          & 88.5          & 84.6          & 90.1          & \multicolumn{1}{c|}{89.3}          & 75.3          & 86.8          & 82.3          & 79.2          & 86.9          & 83.8          & 81.3          & 87.2          & \multicolumn{1}{c|}{85.6}          & 78.5          & 88.1          & 84.4          & 79.8          & 90.7          & 86.6          & 76.0          & 87.9          & \multicolumn{1}{c|}{85.2}          & 84.1                 \\
RetiZero                                                  & 72.1          & 82.5          & 73.2          & 76.2          & 85.0          & 81.9          & 78.7          & 86.5          & \multicolumn{1}{c|}{84.7}          & 58.7          & 76.4          & 68.1          & 60.3          & 78.2          & 71.4          & 61.3          & 79.7          & \multicolumn{1}{c|}{75.5}          & 56.9          & 79.0          & 72.2          & 61.9          & 74.6          & 70.8          & 63.6          & 76.2          & \multicolumn{1}{c|}{72.9}          & 80.1 \\ \hline
\multicolumn{29}{l}{\textit{VLPs with small-scale elite image-text paired data / public categorical data}}    \\ \hdashline
FLAIR                           & 82.9          & 90.1          & 87.9          & \textbf{83.6} & 89.8          & 88.5          & 83.3          & 89.4          & \multicolumn{1}{c|}{88.2}          & 82.2          & 89.7          & 87.5          & 82.5          & 88.6          & 87.1          & 82.7          & 87.1          & \multicolumn{1}{c|}{86.6}          & 82.0          & 90.2          & 88.1          & 82.5          & 91.2          & 88.9          & 82.5          & 91.5          & \multicolumn{1}{c|}{89.5}          & 87.4                 \\
KeepFITV1                                                 & 74.4          & 93.6          & 89.0          & 81.0            & 89.9          & 87.6          & 83.3          & 91.3          & \multicolumn{1}{c|}{89.2}          & 82.2          & 93.5          & 87.9          & 80.0          & 93.1          & 87.9          & 80.8          & 92.8          & \multicolumn{1}{c|}{87.7}          & 81.2          & 93.4          & 87.4          & 83.3          & 94.0          & 89.0          & 81.9          & 92.0          & \multicolumn{1}{c|}{87.6}          & 87.8   \\
\rowcolor[HTML]{EFEFEF}KeepFITV2              & 88.8          & 95.6          & 90.1          & 82.7          & \textbf{91.4} & 87.7          & 85.0          & \textbf{96.0} & \multicolumn{1}{c|}{91.4}          & 86.0          & \textbf{95.0} & 90.3          & 85.8          & \textbf{94.8} & 90.0          & 86.5          & \textbf{94.6} & \multicolumn{1}{c|}{90.5}          & 85.1          & 93.6          & 88.2          & 87.8          & 95.0          & 89.7          & 86.0          & \textbf{95.2} & \multicolumn{1}{c|}{90.5}          & 90.1                  \\
\rowcolor[HTML]{DCDCDC}KeepFITV2$_L$              & \textbf{90.0} & \textbf{96.2} & \textbf{93.4} & 81.1          & 90.3          & \textbf{88.9} & \textbf{91.1} & 94.5          & \multicolumn{1}{c|}{\textbf{94.4}} & \textbf{86.8} & 94.6          & \textbf{91.4} & \textbf{86.2} & 94.4          & \textbf{91.5} & \textbf{87.9} & 94.5          & \multicolumn{1}{c|}{\textbf{91.7}} & \textbf{86.9} & \textbf{95.7} & \textbf{91.4} & \textbf{89.0} & \textbf{95.4} & \textbf{92.2} & \textbf{88.3} & 93.3          & \multicolumn{1}{c|}{\textbf{91.4}} & \textbf{91.5}                 \\ \hline
\multicolumn{29}{c}{ODIR200$\times$3} \\ \hline
\multicolumn{29}{l}{\textit{VLPs with large-scale image-text paired data}}       \\ \hdashline
ViLRef                 & 88.0          & 96.9          & 94.6          & 88.3          & 97.5          & 95.4          & 88.5          & 97.6          & \multicolumn{1}{c|}{95.5}          & 82.2          & 94.3          & 88.1          & 82.3          & 94.4          & 88.0          & 82.3          & 94.5          & \multicolumn{1}{c|}{88.2}          & 82.8          & 93.8          & 87.6          & 80.8          & 93.1          & 85.7          & 82.2          & 93.7          & \multicolumn{1}{c|}{87.0}          & 74.7                 \\
RET-CLIP                                                  & 89.7          & 98.0          & 96.3          & 90.0          & 98.0          & 96.3          & 90.3          & 98.2          & \multicolumn{1}{c|}{96.8}          & 83.3          & 94.4          & 90.2          & 84.5          & 95.8          & 92.9          & 85.7          & 96.7          & \multicolumn{1}{c|}{94.7}          & 82.8          & 94.6          & 90.8          & 85.8          & 96.5          & 93.9          & 88.8          & 97.1          & \multicolumn{1}{c|}{95.0}          & 84.9                 \\
{RetiZero}                           & 89.7          & 98.1          & 96.5          & 92.2          & 98.3          & 96.8          & 91.5          & 98.4          & \multicolumn{1}{c|}{97.0}          & 71.8          & 96.7          & 93.7          & 73.0          & 97.0          & 94.1          & 75.5          & 97.2          & \multicolumn{1}{c|}{94.6}          & 71.7          & 96.4          & 93.0          & 76.3          & 96.8          & 93.7          & 80.0          & 97.6          & \multicolumn{1}{c|}{95.1}          & 95.4                 \\ \hline
\multicolumn{29}{l}{\textit{VLPs with small-scale elite image-text paired data / public categorical data}}    \\ \hdashline
FLAIR                  & 72.0          & 89.4          & 83.2          & 83.2          & 92.9          & 88.4          & 87.0          & 95.7          & \multicolumn{1}{c|}{92.4}          & 39.7          & 83.7          & 69.9          & 41.2          & 84.4          & 71.3          & 42.7          & 85.1          & \multicolumn{1}{c|}{72.4}          & 40.7          & 83.9          & 70.0          & 45.5          & 85.7          & 73.3          & 53.8          & 88.0          & \multicolumn{1}{c|}{77.5}          & 73.8                 \\
KeepFITV1                                                 & 84.2          & \textbf{96.3} & \textbf{93.9} & 87.7          & \textbf{96.9} & 94.9          & \textbf{89.7} & \textbf{97.4} & \multicolumn{1}{c|}{95.5}          & 81.7          & 93.4          & 88.2          & 83.3          & 94.2          & 89.6          & 84.2          & 95.0          & \multicolumn{1}{c|}{91.0}          & 81.3          & 93.9          & 89.3          & 84.2          & 94.8          & 90.8          & 86.3          & 95.9          & \multicolumn{1}{c|}{92.8}          & 90.6                 \\
\rowcolor[HTML]{EFEFEF}KeepFITV2              & 81.2          & 94.3          & 90.7          & 86.5          & 95.8          & 93.0          & 87.7          & 96.5          & \multicolumn{1}{c|}{94.2}          & \textbf{84.3} & 94.2          & \textbf{90.2} & 86.0          & 95.1          & 91.7          & 86.3 & 95.6          & \multicolumn{1}{c|}{92.8}          & 84.2          & 94.0          & \textbf{89.9} & 85.1          & 95.1          & 91.8          & 87.0          & 95.7          & \multicolumn{1}{c|}{92.9}          & 90.8                 \\
\rowcolor[HTML]{DCDCDC}KeepFITV2$_L$              & \textbf{85.0} & 95.9          & 93.2          & \textbf{87.5} & \textbf{96.9} & \textbf{94.8} & 89.3          & \textbf{97.4} & \multicolumn{1}{c|}{\textbf{95.4}} & 84.2          & \textbf{94.3} & 90.1          & \textbf{87.2} & \textbf{95.2} & \textbf{91.9} & \textbf{87.7}          & \textbf{95.8} & \multicolumn{1}{c|}{\textbf{92.9}} & \textbf{85.0} & \textbf{94.5} & 89.7          & \textbf{87.3} & \textbf{95.6} & \textbf{92.7} & \textbf{87.8} & \textbf{96.2} & \multicolumn{1}{c|}{\textbf{93.3}} & \textbf{91.7}                 \\ \hline
\multicolumn{29}{c}{Retina}  \\ \hline
\multicolumn{29}{l}{\textit{VLPs with large-scale image-text paired data}}   \\ \hdashline
ViLRef                 & 61.2          & 84.9          & 70.2          & 61.9          & 84.8          & 68.4          & 66.0          & 86.3          & \multicolumn{1}{c|}{72.3}          & 50.2          & 77.4          & 58.0          & 50.7          & 77.6          & 58.2          & 50.9          & 77.8          & \multicolumn{1}{c|}{58.6}          & 52.0          & 76.1          & 56.0          & 52.7          & 77.0          & 56.6          & 52.2          & 77.3          & \multicolumn{1}{c|}{57.3}          & 65.7                 \\
RET-CLIP                                                  & 65.2          & 88.3          & 75.3          & 65.7          & 86.3          & 72.1          & 67.4          & 87.1          & \multicolumn{1}{c|}{72.6}          & 57.9          & 83.4          & 66.3          & 57.3          & 83.9          & 66.5          & 58.2          & 84.7          & \multicolumn{1}{c|}{67.9}          & 57.0          & 83.6          & 66.6          & 59.6          & 83.9          & 67.2          & 65.5          & 84.8          & \multicolumn{1}{c|}{68.6}          & 72.0                 \\
RetiZero                                                  & 58.4          & 83.7          & 67.2          & 59.4          & 82.0          & 64.4          & 62.6          & 84.4          & \multicolumn{1}{c|}{68.0}          & 41.3          & 75.7          & 54.4          & 42.0          & 75.7          & 54.7          & 42.9          & 75.9          & \multicolumn{1}{c|}{55.5}          & 41.9          & 77.0          & 55.7          & 41.3          & 76.6          & 54.6          & 45.9          & 77.9          & \multicolumn{1}{c|}{58.3}          & 70.0                 \\ \hline
\multicolumn{29}{l}{\textit{VLPs with small-scale elite image-text paired data / public categorical data}}     \\ \hdashline
FLAIR                  & 41.1          & 66.8          & 46.6          & 42.9          & 71.2          & 50.0          & 53.6          & 78.8          & \multicolumn{1}{c|}{59.0}          & 33.9          & 67.3          & 45.2          & 34.1          & 67.5          & 45.3          & 34.3          & 67.8          & \multicolumn{1}{c|}{45.6}          & 35.8          & 66.3          & 45.2          & 36.4          & 66.3          & 43.8          & 39.2          & 68.5          & \multicolumn{1}{c|}{47.5}          & 51.9                 \\
KeepFITV1                                                 & 57.6          & 82.1          & 65.2          & \textbf{62.7} & \textbf{83.1} & \textbf{69.3} & \textbf{66.0}          & \textbf{85.5}          & \multicolumn{1}{c|}{\textbf{72.7}} & 41.8          & 76.3          & 54.4          & 42.3          & 77.2          & 55.4          & 43.4          & 78.1          & \multicolumn{1}{c|}{56.8}          & 41.9          & 77.0          & 55.6          & 43.3          & 79.3          & 58.2          & 45.2          & 78.9          & \multicolumn{1}{c|}{59.4}          & 63.3                 \\
\rowcolor[HTML]{EFEFEF}KeepFITV2              & \textbf{59.2} & \textbf{82.2} & \textbf{66.1} & 56.4          & 80.7          & 61.7          & \textbf{64.7} & 85.2 & \multicolumn{1}{c|}{70.0}          & 43.4          & \textbf{79.3} & \textbf{59.1} & 44.7          & \textbf{79.6} & \textbf{59.7} & 44.7          & \textbf{80.4} & \multicolumn{1}{c|}{\textbf{61.1}} & 43.2          & \textbf{79.9} & \textbf{59.5} & 43.4          & 79.9          & 61.6          & 47.8          & \textbf{81.5} & \multicolumn{1}{c|}{63.3}          & 64.4                 \\
\rowcolor[HTML]{DCDCDC}KeepFITV2$_L$              & 58.2          & 81.2          & 65.9          & 58.0          & 81.0          & 66.0          & 62.4          & 82.2          & \multicolumn{1}{c|}{66.4}          & \textbf{44.8} & 77.0          & 57.8          & \textbf{45.8} & 78.3          & 59.6          & \textbf{46.1} & 78.8          & \multicolumn{1}{c|}{60.8}          & \textbf{44.9} & 77.6          & 59.1          & \textbf{49.5} & \textbf{81.6} & \textbf{63.6} & \textbf{51.0} & 80.2          & \multicolumn{1}{c|}{\textbf{64.0}} & \textbf{64.5}                 \\ \hline
\multicolumn{29}{c}{iChallenge-AMD}     \\ \hline
\multicolumn{29}{l}{\textit{VLPs with large-scale image-text paired data}}  \\ \hdashline
ViLRef                  & 83.9          & 94.1          & 92.6          & 83.0          & 93.4          & 92.3          & 89.2          & 95.1          & \multicolumn{1}{c|}{93.5}          & 76.9          & 88.3          & 84.8          & 77.5          & 88.3          & 84.8          & 77.5          & 88.4          & \multicolumn{1}{c|}{84.8}          & 78.1          & 87.1          & 83.4          & 77.9          & 87.8          & 84.4          & 79.6          & 86.4          & \multicolumn{1}{c|}{83.1}          & 85.8                 \\
RET-CLIP                                                  & 82.4          & 94.0          & 82.8          & 84.1          & 94.0          & 92.6          & 85.9          & 95.7          & \multicolumn{1}{c|}{94.1}          & 80.7          & 88.6          & 87.0          & 79.7          & 88.6          & 87.0          & 77.7          & 89.0          & \multicolumn{1}{c|}{87.3}          & 77.9          & 87.9          & 86.3          & 80.1          & 90.1          & 89.4          & 81.2          & 95.9          & \multicolumn{1}{c|}{90.6}          & 87.1                 \\
{RetiZero}                           & 81.6          & 87.5          & 86.9          & 79.7          & 89.5          & 88.2          & 81.9          & 92.2          & \multicolumn{1}{c|}{90.9}          & 66.9          & 81.4          & 79.5          & 68.1          & 81.8          & 79.7          & 68.4          & 82.5          & \multicolumn{1}{c|}{80.3}          & 69.1          & 84.1          & 82.3          & 68.8          & 81.1          & 79.2          & 70.1          & 87.2          & \multicolumn{1}{c|}{85.0}          & 86.5                 \\ \hline
\multicolumn{29}{l}{\textit{VLPs with small-scale elite image-text paired data / public categorical data}}     \\ \hdashline
FLAIR                  & 65.3          & 79.1          & 75.1          & 77.5          & 85.6          & 81.8          & 81.3          & 89.8          & \multicolumn{1}{c|}{85.9}          & 68.9          & 80.1          & 75.0          & 68.9          & 80.7          & 75.6          & 69.1          & 81.5          & \multicolumn{1}{c|}{76.2}          & 70.7          & 79.9          & 74.3          & 69.2          & 79.7          & 74.3          & 73.9          & 84.6          & \multicolumn{1}{c|}{79.6}          & 77.1                 \\
KeepFITV1                                                 & 69.4          & 88.6          & 86.9          & 78.8          & 91.3          & 88.5          & 79.4          & 92.3          & \multicolumn{1}{c|}{89.7}          & 76.0          & 87.9          & 86.4          & 76.1          & 88.6          & 86.9          & 75.5          & \textbf{89.7} & \multicolumn{1}{c|}{87.8}          & 75.7          & 89.3          & 87.7          & 76.4          & \textbf{89.6} & 87.7          & 78.9          & 90.6          & \multicolumn{1}{c|}{88.8}          & 84.6                 \\
\rowcolor[HTML]{EFEFEF}KeepFITV2              & 78.2          & 88.9          & 86.9          & \textbf{81.6} & \textbf{91.7} & 88.7          & \textbf{84.0} & 92.2          & \multicolumn{1}{c|}{88.8}          & 80.9          & 88.0          & 87.4          & 82.1          & 87.5          & 86.5          & 81.5          & 87.9          & \multicolumn{1}{c|}{86.9}          & \textbf{81.5} & 88.7          & 87.3          & \textbf{81.9} & 87.8          & 86.4          & \textbf{82.0} & 89.7          & \multicolumn{1}{c|}{87.9}          & 86.0                 \\
\rowcolor[HTML]{DCDCDC}KeepFITV2$_L$              & \textbf{80.7} & \textbf{90.4} & \textbf{89.6} & 79.3          & 89.7          & \textbf{88.8} & 82.7          & \textbf{92.9} & \multicolumn{1}{c|}{\textbf{90.7}} & \textbf{82.6} & \textbf{89.2} & \textbf{88.8} & \textbf{82.4} & \textbf{89.1} & \textbf{88.7} & \textbf{82.0} & 89.2          & \multicolumn{1}{c|}{\textbf{88.7}} & 80.7          & \textbf{89.9} & \textbf{88.8} & 79.8          & 88.7          & \textbf{87.9} & 81.7          & \textbf{90.8} & \multicolumn{1}{c|}{\textbf{89.5}} & \textbf{86.8}                 \\ \hline
\end{tabular}
\end{table*}

\subsection{Datasets}
\subsubsection{CFP Modality}
KeepFIT V2 in CFP modality is trained on the proposed image-text MM-Retinal V2 and the categorical public retinal datasets from flair\cite{silva2023foundation} which consist of over 190K images across 96 categories. 
Besides, we additionally collect over 80K data samples from public categorical retinal datasets. Using all 270K public datasets along with MM-Retinal V2, we trained KeepFIT V2$_L$.
For evaluation, we utilize REFUGE\cite{orlando2020refuge}, ODIR200$\times$3\cite{ODIR2019}, iChallenge-AMD\cite{ADAM}, Retina\cite{cataract2020}, FIVES\cite{jin2022fives} and APTOS\cite{APTOS2019} to perform various downstream classification tasks. These evaluation datasets encompass several common fundus diseases, including glaucoma, pathologic myopia, cataract, diabetic retinopathy, age-related macular degeneration, and other retinal disorders.

\subsubsection{FFA Modality}
The pretraining for FFA modality is conducted on MM-Retinal V2 and FFA-IR\cite{li2021ffa}, an image-text paired dataset comprising 10,790 reports and 1,048,584 FFA images, spanning 46 retinal lesion categories. 
We extensively evaluate KeepFIT V2 using two FFA public datasets. MPOS\cite{wang2024non} includes 600 images across four fundus disease categories. AngioReport (APTOS2023)\cite{zhang2023angiographic} contains a total of over 50K images, covering 24 distinct categories.

\subsubsection{OCT Modality}
For pretraining, besides MM-Retinal V2, we also collect eleven OCT public datasets with only category labels for KeepFIT V2 pretraining, totaling over 181K images. 
In addition, OCTID\cite{gholami2020octid} and OCTDL\cite{kulyabin2024octdl} datasets are used for evaluation, consisting of 9 common fundus diseases.

\subsection{Methods for Comparison}
For CFP modality, we executed the downstream tasks across six methods to ensure a comprehensive comparison. Specifically, the methods for comparison can be divided into two groups. The first group is VLPs with categorical public datasets/small-scale image-text paired data. This group includes models of FLAIR\cite{silva2023foundation}, KeepFIT V1\cite{mm}, KeepFIT V2, and KeppFIT V$_L$. In contrast, the second group is VLPs with large-scale private image-text paired data, including RET-CLIP\cite{du2024ret}, ViLReF\cite{yang2024vilref}, and RetiZero\cite{wang2024common}. These models are trained on 193,865, 451,956, and 341,896 image-text pairs, respectively. Compared to RET-CLIP, ViLReF, and RetiZero, KeepFIT V1 and V2 utilize only 1\% of the image-text paired data.
Moreover, RETFound\cite{zhou2023foundation} is an exception. It is trained using a masked image modeling (MIM) approach. Thus, owing to the absence of text encoder, RETFound can only be evaluated in linear probing setting. 

Due to the scarcity of released available comparable models for FFA and OCT modalities, we selected two generalist vision-language models in the medical domain for biomedical understanding: BiomedCLIP\cite{zhang2023biomedclip} and PubMedCLIP\cite{eslami2023pubmedclip}. BiomedCLIP is trained on 15M biomedical image-text pairs from 4.4M scientific articles in PMC. PubMedCLIP is pre-trained on ROCO dataset\cite{pelka2018radiology}, which comprises over 80K samples spanning a wide range of medical imaging modalities, including ultrasound, X-rays, MRI, angiography, etc. Additionally, we also incorporate CLIP as a comparison model, training on the same retinal public datasets as our models, rather than using its original pretrained weights.

\subsection{Implementation Details}
Following the design in KeepFIT V1, we adopt the same model architecture of image and text encoders and hyperparameter settings. Specifically, all the images are resized to a resolution of 512$\times$512 and the texts are in English version with a maximum length of 256 tokens. The multi-head cross-attention modules are trained from scratch with a feature dimension of 512. Our model is trained with the AdamW optimizer with a learning rate of 1$\times$10$^{-4}$ and a weight decay of 1$\times$10$^{-5}$. A cosine scheduler is used with a warm-up for the first epoch. The visual tokenizer is trained using 16,384 codebook size and 256 code dimension as the default setting in \cite{shi2024taming}. All the metrics presented are averaged across five cross-validation folds.

\subsection{Evaluation of the CFP Modality}

\begin{figure}[!t]
\centerline{\includegraphics[width=\columnwidth]{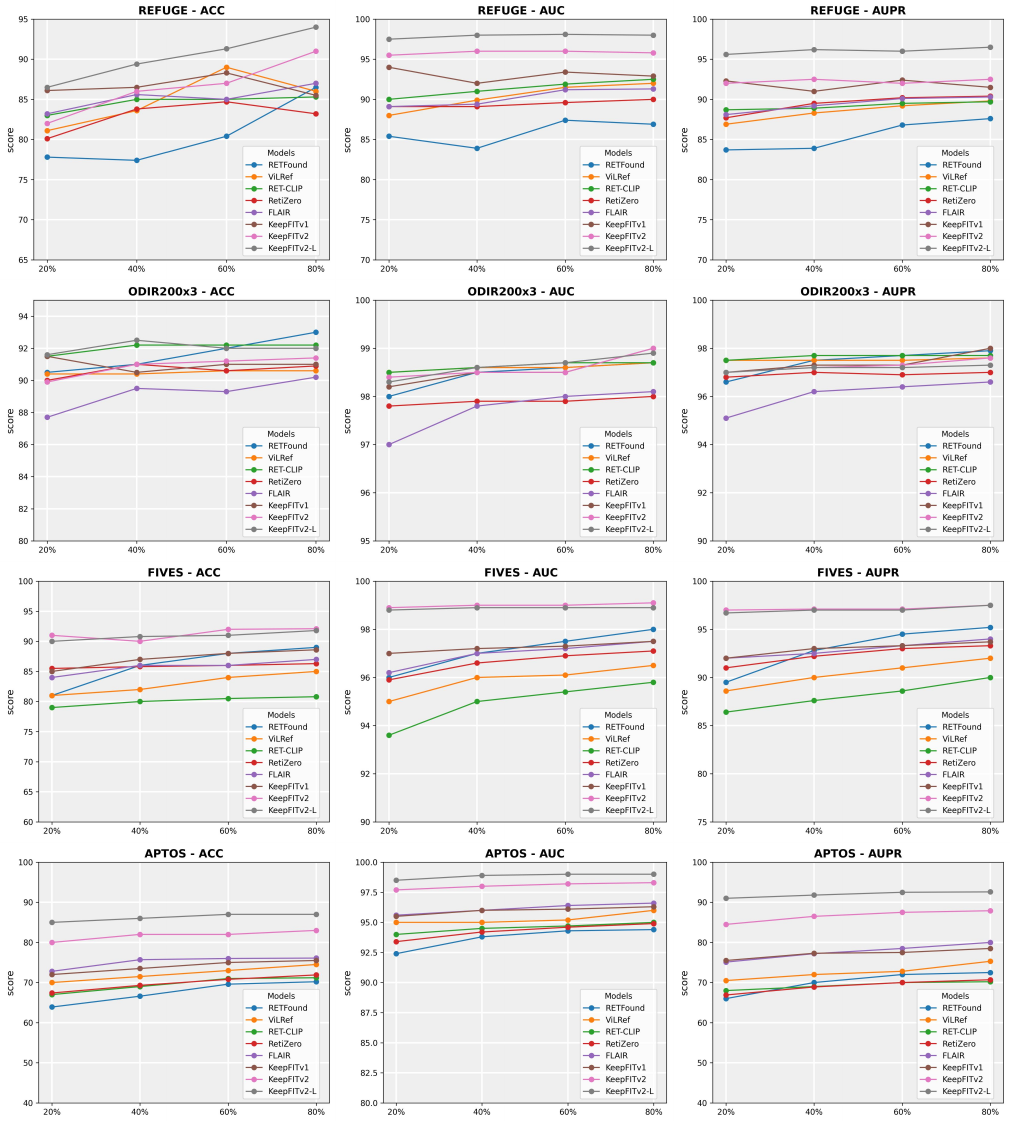}}
% \centerline{\includegraphics[scale=0.8]{assets/CFP_lp.pdf}}
\caption{Linear probing classification performance in CFP modality across datasets (each row represents the results of a dataset).}
\label{CFP_lp}
\end{figure}

\subsubsection{Zero-shot}
Table~\ref{CFP_zeroshot} reports the zero-shot experimental results on four unseen downstream datasets in terms of ACC, AUC, and AUPR. In addition, ODIR200$\times$3 includes two unseen categories during training, which are pathologic myopia and cataract. From an overall perspective, KeepFIT V2 and KeepFIT V2$_L$ surpass the KeepFIT V1 on all the downstream datasets, achieving an average score improvement of up to 3.4\%, 2.9\%, 3.7\%, and 4.2\% on REFUGE, ODIR200$\times$3, Retina, and iChallenge-AMD, respectively. 
With the expansion of training data volume, though KeepFIT V2$_L$ shows a slight performance decline on REFUGE and Retina, it achieves an improvement of 2.9\% on ODIR200$\times$3 and 2.1\% on iChallenge-AMD when compared to KeepFIT V2, indicating that a larger volume of training data can lead to performance gains.

It is worth noting that ViLRef, RET-CLIP, and RetiZero were trained on private datasets of 200K to 450K image-text pairs, which have not been publicly released. This may explain the performance gap between VLP models trained on large-scale and small-scale image-text paired data. Nonetheless, the KeepFIT V2 model consistently outperforms all comparison models on the REFUGE dataset, achieving improvements of 22.6\%, 4.3\%, and 24.0\%, respectively. 
These results highlight that, even with a modest high-quality image-text dataset, KeepFIT V2 could effectively incorporate expert knowledge from MM-Retinal V2 into foundational vision-language pertaining, by spreading such an elite spark on public retinal datasets, demonstrating its exceptional generalization capabilities.

\begin{table}[]
\caption{Zero-Shot Classification Performance in FFA Modality across Two Datasets. (\%)}
\label{FFA_zeroshot}
\setlength{\tabcolsep}{3.7pt}
\renewcommand\arraystretch{1.2}
\begin{tabular}{ccccccccc}
\hline
\multirow{2}{*}{Model} & \multicolumn{4}{c|}{Angiographic}                  & \multicolumn{4}{c}{MPOS}     \\ \cline{2-9} 
                       & ACC   & AUC   & AUPR  & \multicolumn{1}{c|}{AVG}   & ACC   & AUC   & AUPR  & AVG   \\ \hline
\multicolumn{9}{l}{\textit{Generalist VLPs for biomedical understanding}}         \\ \hdashline
BiomedCLIP             & 4.7   & 49.9  & 4.7   & \multicolumn{1}{c|}{19.8}  & 20.6  & 65.6  & 34.5 & 40.2       \\
PubMedCLIP             & 4.7   & 50.5  & 4.8   & \multicolumn{1}{c|}{20.0}  & 18.5  & 48.8  & 19.8 & 29.0       \\ \hline
\multicolumn{9}{l}{\textit{Specialist VLPs for retinal understanding}}      \\ \hdashline
CLIP                   & 14.2  & 59.5  & 5.7   & \multicolumn{1}{c|}{26.5}  & 31.5  & 62.8  & 32.3 & 42.2 \\
KeepFITV1              & 11.3  & 62.0  & 6.1   & \multicolumn{1}{c|}{26.5}  & 31.2  & 64.7  & 33.4 & 43.1 \\
\rowcolor[HTML]{DCDCDC}KeepFITV2               & \textbf{15.1} & \textbf{69.0} & \textbf{9.1}  & \multicolumn{1}{c|}{\textbf{31.1}} & \textbf{64.8} & \textbf{89.7} & \textbf{73.8} & \textbf{76.1} \\ \hline
\end{tabular}
\end{table}

\subsubsection{Few-Shot}
Next, we assess the performance of the proposed KeepFIT V2 in low-data regimes by conducting few-shot classification experiments. These experiments are derived by varying the number of shots (images per category) used for adaptation with the utility of Clip-Adapter \cite{gao2024clip} and Tip-Adapter \cite{zhang2022tip}. 
From Table~\ref{CFP_fewshot}, KeepFIT V2 and KeepFIT V2$_L$ consistently outperform KeepFIT V1, improving the average score of ACC, AUC and AUPR metrics with a maximum improvement of 3.7\% on REFUGE, 1.1\% on ODIR 200$\times$3, 1.2\% on Retina, and 2.2\% on iChallenge-AMD when compared to KeepFIT V1.

Our KeepFIT V2 and KeepFIT V2$_L$ achieve top-2 performance on the REFUGE and iChallenge-AMD datasets, as well as top-3 performance on the ODIR200$\times$3 dataset. Notably, they not only outperform FLAIR and KeepFIT V1 but also surpass models that are trained on large-scale image-text paired datasets. These results point to the superior representation and vision-language alignment capability of KeepFIT V2, underscoring its generalizability under limited image-text data resource. 

\subsubsection{Linear Probing}
To further evaluate the effectiveness and transferability of the proposed KeepFIT V2, we conduct linear probing experiments. Concretely, we use the frozen image encoder from KeepFIT V2 as image feature extractor and incorporate an additional linear layer as the classifier, whose parameters are fine-tuned on downstream datasets. 

The experiment results are presented in Fig.~\ref{CFP_lp}. Compared to VLPs trained on small-scale elite image-text paired data / public categorical data (i.e. FLAIR, RETFound, and KeepFIT V1), our KeepFIT V2 and KeepFIT V2$_L$ achieve significant performance improvement across all metrics. Furthermore, when compared to VLPs trained on large-scale image-text paired data (i.e., ViLRef, RET-CLIP, and RetiZero), KeepFIT V2 and KeepFIT V2$_L$ still maintain comparable or even superior performance. From the above experiments in CFP modality, we can conclude that no foundation models can perform the best across all datasets in every evaluation setting. This indicates that building a retinal foundation model that performs well across diverse scenarios and various disease types is highly challenging and calls for further exploration.

\subsection{Evaluation of the FFA Modality}

\begin{table*}[]
\caption{Few-Shot Classification Performance in FFA Modality across Two Datasets. (\%)}
\label{FFA_fewshot}
\setlength{\tabcolsep}{0.5pt}
\renewcommand\arraystretch{1.2}
\begin{tabular}{ccccccccccccccccccccccccccccc}
\hline
\multirow{3}{*}{Model} & \multicolumn{9}{c|}{Clipadapter}                                                                                                                                   & \multicolumn{9}{c|}{Tipadapter}                                                                                                                                    & \multicolumn{9}{c|}{Tipadapter-f}                                                                                                                                  & \multirow{3}{*}{AVG} \\
                       & \multicolumn{3}{c}{1}                         & \multicolumn{3}{c}{5}                         & \multicolumn{3}{c|}{10}                                            & \multicolumn{3}{c}{1}                         & \multicolumn{3}{c}{5}                         & \multicolumn{3}{c|}{10}                                            & \multicolumn{3}{c}{1}                         & \multicolumn{3}{c}{5}                         & \multicolumn{3}{c|}{10}                                            &                      \\ \cline{2-28}
                       & {\fontsize{7pt}{12pt}\selectfont ACC}   & {\fontsize{7pt}{12pt}\selectfont AUC}   & {\fontsize{7pt}{12pt}\selectfont AUPR}  & {\fontsize{7pt}{12pt}\selectfont ACC}   & {\fontsize{7pt}{12pt}\selectfont AUC}   & {\fontsize{7pt}{12pt}\selectfont AUPR}  & {\fontsize{7pt}{12pt}\selectfont ACC}   & {\fontsize{7pt}{12pt}\selectfont AUC}   & \multicolumn{1}{c|}{{\fontsize{7pt}{12pt}\selectfont AUPR}}  & {\fontsize{7pt}{12pt}\selectfont ACC}   & {\fontsize{7pt}{12pt}\selectfont AUC}   & {\fontsize{7pt}{12pt}\selectfont AUPR}  & {\fontsize{7pt}{12pt}\selectfont ACC}   & {\fontsize{7pt}{12pt}\selectfont AUC}   & {\fontsize{7pt}{12pt}\selectfont AUPR}  & {\fontsize{7pt}{12pt}\selectfont ACC}   & {\fontsize{7pt}{12pt}\selectfont AUC}   & \multicolumn{1}{c|}{{\fontsize{7pt}{12pt}\selectfont AUPR}}  & {\fontsize{7pt}{12pt}\selectfont ACC}   & {\fontsize{7pt}{12pt}\selectfont AUC}   & {\fontsize{7pt}{12pt}\selectfont AUPR}  & {\fontsize{7pt}{12pt}\selectfont ACC}   & {\fontsize{7pt}{12pt}\selectfont AUC}   & {\fontsize{7pt}{12pt}\selectfont AUPR}  & {\fontsize{7pt}{12pt}\selectfont ACC}   & {\fontsize{7pt}{12pt}\selectfont AUC}   & \multicolumn{1}{c|}{{\fontsize{7pt}{12pt}\selectfont AUPR}}  &                               \\ \hline
\multicolumn{29}{c}{Angiographic}                      \\ \hline
\multicolumn{29}{l}{\textit{Generalist VLPs for biomedical understanding}}            \\ \hdashline
BiomedCLIP             & 12.3          & 63.0          & 9.8           & 25.3          & 73.2          & 16.4          & 34.6          & 78.6          & \multicolumn{1}{c|}{22.2}          & 5.6           & 56.2          & 6.5           & 6.7           & 58.7          & 6.8           & 7.3           & 61.7          & \multicolumn{1}{c|}{7.2}           & 5.4           & 56.5          & 6.6           & 8.9           & 60.0          & 7.7           & 15.0          & 66.1          & \multicolumn{1}{c|}{9.9}           & 29.2                 \\
PubMedCLIP             & 15.7          & 60.7          & 10.6          & 27.4          & 70.9          & 17.1          & 35.8          & 75.7          & \multicolumn{1}{c|}{21.6}          & 5.5           & 52.7          & 5.6           & 6.1           & 60.2          & 6.3           & 9.2           & 63.9          & \multicolumn{1}{c|}{8.4}           & 5.6           & 52.3          & 5.7           & 9.3           & 61.8          & 8.7           & 12.2          & 67.1          & \multicolumn{1}{c|}{10.3}          & 29.1                 \\ \hline
\multicolumn{29}{l}{\textit{Specialist VLPs for retinal understanding}}    \\ \hdashline
CLIP                   & 20.6          & 67.4          & 13.3          & 41.8          & 79.0          & 30.7          & 53.2          & 84.1          & \multicolumn{1}{c|}{42.0}          & 7.2           & 59.2          & 6.3           & 7.6           & 60.1          & 6.5           & 8.2           & 61.1          & \multicolumn{1}{c|}{6.9}           & 7.3           & 59.3          & 6.5           & 9.0           & 62.7          & 7.7           & 15.4          & 67.1          & \multicolumn{1}{c|}{10.5}          & 33.4                 \\
KeepFITV1               & 24.0          & 70.8          & 16.4          & \textbf{42.2} & 79.0          & \textbf{31.8} & \textbf{51.9} & \textbf{85.2} & \multicolumn{1}{c|}{\textbf{41.9}} & 10.9          & 64.6          & 6.8           & 13.1          & 66.4          & 7.6           & 14.4          & 68.2          & \multicolumn{1}{c|}{9.8}           & 10.9          & 65.0          & 6.9           & 17.4          & 69.8          & 12.6          & 29.4          & 76.4          & \multicolumn{1}{c|}{22.2}          & 37.6                 \\
\rowcolor[HTML]{DCDCDC}KeepFITV2              & \textbf{24.1} & \textbf{72.8} & \textbf{16.5} & 39.1          & \textbf{79.2}          & 27.0          & 48.2          & 83.9          & \multicolumn{1}{c|}{35.6}          & \textbf{13.2} & \textbf{71.3} & \textbf{10.3} & \textbf{17.5} & \textbf{73.2} & \textbf{12.9} & \textbf{22.5} & \textbf{75.3} & \multicolumn{1}{c|}{\textbf{16.6}} & \textbf{13.5} & \textbf{71.7} & \textbf{10.3} & \textbf{25.9} & \textbf{76.8} & \textbf{18.5} & \textbf{39.4} & \textbf{82.4} & \multicolumn{1}{c|}{\textbf{30.4}} & \textbf{41.0}        \\ \hline
\multicolumn{29}{c}{MPOS}     \\ \hline
\multicolumn{29}{l}{\textit{Generalist VLPs for biomedical understanding}}    \\ \hdashline
BiomedCLIP             & 38.7          & 68.5          & 41.7          & 53.0          & 78.3          & 53.0          & 58.2          & 83.3          & \multicolumn{1}{c|}{62.3}          & 21.1          & 62.9          & 30.7          & 21.1          & 64.1          & 31.6          & 21.5          & 65.5          & \multicolumn{1}{c|}{33.0}          & 20.5          & 59.1          & 28.4          & 21.6          & 63.4          & 31.1          & 24.4          & 67.0          & \multicolumn{1}{c|}{37}            & 46.0                 \\
PubMedCLIP             & 28.1          & 57.8          & 31.2          & 42.9          & 72.0          & 43.7          & 49.4          & 77.0          & \multicolumn{1}{c|}{52.0}          & 16.4          & 47.1          & 20.4          & 17.7          & 50.0          & 22.5          & 21.7          & 54.2          & \multicolumn{1}{c|}{25.7}          & 17.1          & 49.3          & 21.6          & 24.0          & 56.2          & 27.7          & 24.7          & 61.8          & \multicolumn{1}{c|}{32.5}          & 38.7                 \\ \hline
\multicolumn{29}{l}{\textit{Specialist VLPs for retinal understanding}}    \\ \hdashline
CLIP                   & 46.0          & 75.1          & 47.6          & 66.7          & 90.4          & 75.1          & 77.1          & 94.3          & \multicolumn{1}{c|}{83.7}          & 30.3          & 61.1          & 32.5          & 31.0          & 62.0          & 33.0          & 31.7          & 63.1          & \multicolumn{1}{c|}{34.3}          & 28.2          & 60.7          & 32.5          & 29.9          & 62.8          & 33.9          & 36.3          & 67.9          & \multicolumn{1}{c|}{38.1}          & 52.8                 \\
KeepFITV1              & 58.0          & 85.3          & 64.4          & \textbf{80.2} & 94.1          & \textbf{87.3} & \textbf{85.8} & \textbf{95.8} & \multicolumn{1}{c|}{\textbf{92.8}} & 32.7          & 64.8          & 35.5          & 39.0          & 69.9          & 42.6          & 43.5          & 75.7          & \multicolumn{1}{c|}{50.7}          & 33.7          & 65.8          & 36.7          & 44.5          & 73.8          & 49.5          & 62.3          & 85.2          & \multicolumn{1}{c|}{67.6}          & 63.6                 \\
\rowcolor[HTML]{DCDCDC}KeepFITV2              & \textbf{73.4} & \textbf{92.6} & \textbf{80.4} & 78.3          & \textbf{94.3} & 84.8 & 82.2          & \textbf{95.8}          & \multicolumn{1}{c|}{88.5} & \textbf{68.4} & \textbf{90.3} & \textbf{75.3} & \textbf{69.4} & \textbf{91.4} & \textbf{78.2} & \textbf{72.0} & \textbf{92.4} & \multicolumn{1}{c|}{\textbf{81.2}} & \textbf{66.6} & \textbf{90.2} & \textbf{75.6} & \textbf{69.5} & \textbf{91.9} & \textbf{80.2} & \textbf{75.4} & \textbf{93.6} & \multicolumn{1}{c|}{\textbf{84.5}} & \textbf{82.1}        \\ \hline
\end{tabular}
\end{table*}

\begin{figure}[!t]
\centerline{\includegraphics[width=\columnwidth]{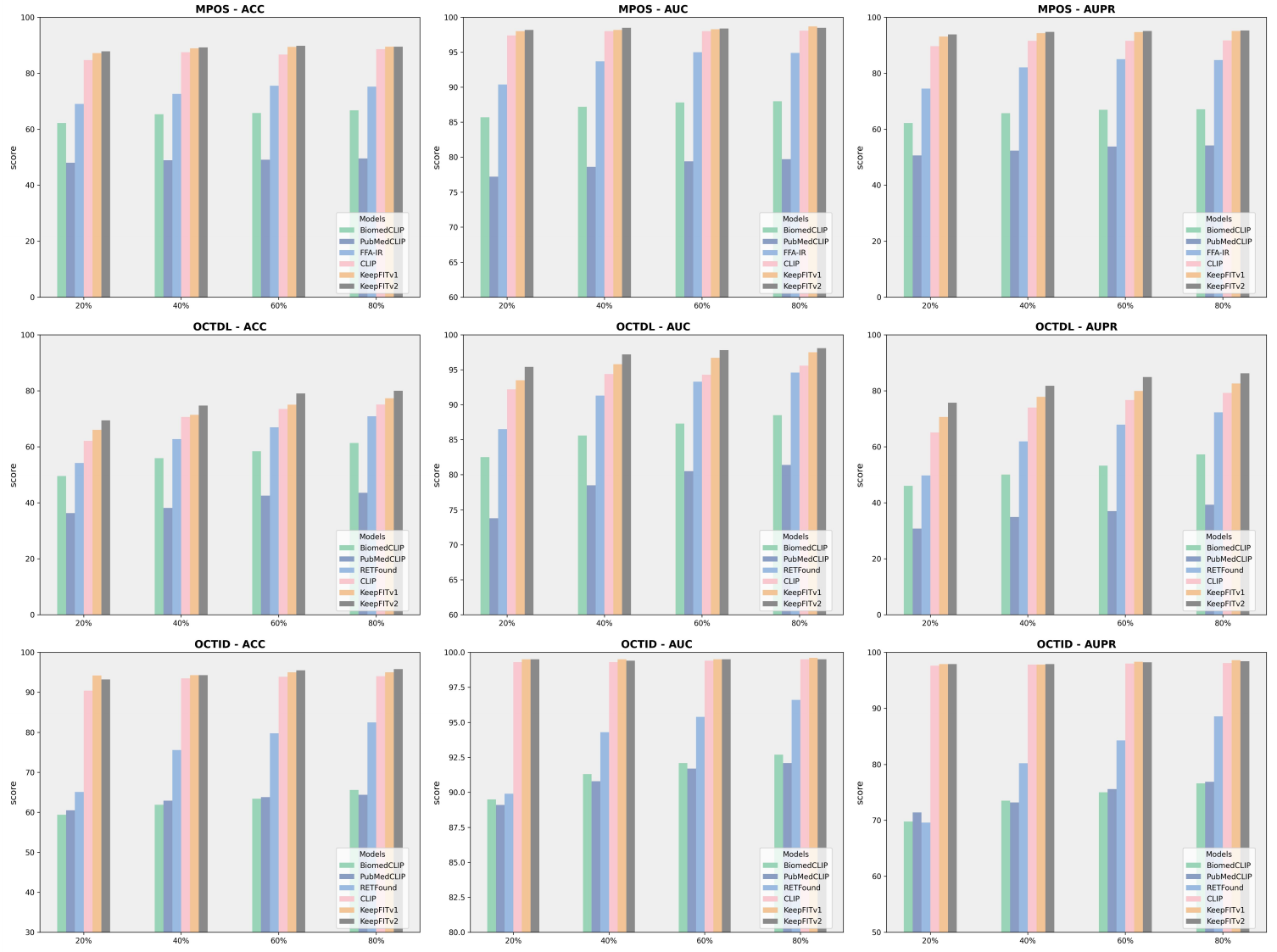}}
% \centerline{\includegraphics[scale=0.8]{assets/OCT_lp.pdf}}
\caption{Linear probing classification performance in FFA and OCT modalities across datasets (each row represents the results of a dataset).}
\label{FFA_OCT_lp}
\end{figure}

This subsection evaluates the performance of KeepFIT V2 across zero-shot, few-shot, and linear probing settings in FFA modality. Table~\ref{FFA_zeroshot}, Table~\ref{FFA_fewshot}, and Fig.~\ref{FFA_OCT_lp} show that KeepFIT V2 achieves substantial improvements over all the compared methods. Under the zero-shot classification setting, it achieves overall performance increases of 4.6\% and 33.0\% for the Angiographic and MPOS datasets, respectively. Additionally, under the few-shot setting, it clearly outperforms the previous KeepFIT V1 by 3.4\% and 18.5\% for the Angiographic and MPOS datasets, respectively.

The performance improvement from KeepFIT V1 to KeepFIT V2 indicates that in the FFA modality, our hybrid image-text knowledge injection module, particularly the appearance-oriented component, effectively utilizes the detailed features learned by the image tokenizer to achieve precise image retrieval between MM-Retinal V2 and public datasets. This enables the effective injection of expert knowledge from the elite MM-Retinal V2 into the pretraining process of KeepFIT V2. Furthermore, it underscores the essential role of detailed features in the understanding and analysis of FFA images.  

\subsection{Evaluation of the OCT Modality}
\begin{table}[]
\caption{Zero-Shot Classification Performance in OCT Modality across Two datasets. (\%)}
\label{OCT_zeroshot}
\setlength{\tabcolsep}{3.7pt}
\renewcommand\arraystretch{1.2}
\begin{tabular}{ccccccccc}
\hline
\multirow{2}{*}{Model} & \multicolumn{4}{c|}{OCTDL}                         & \multicolumn{4}{c}{OCTID}     \\ \cline{2-9} 
                       & ACC   & AUC   & AUPR  & \multicolumn{1}{c|}{AVG}   & ACC   & AUC   & AUPR  & AVG   \\ \hline
\multicolumn{9}{l}{\textit{Generalist VLPs for biomedical understanding}}         \\ \hdashline
BiomedCLIP                  & 21.3          & 61.6          & 21.7          & \multicolumn{1}{c|}{34.9}          & 20.8          & 68.4          & 43.5          & 44.2          \\
PubMedCLIP                                            & 14.9          & 51.0          & 16.2          & \multicolumn{1}{c|}{27.4}          & 11.9          & 48.2          & 21.1          & 27.1          \\\hline
\multicolumn{9}{l}{\textit{Specialist VLPs for retinal understanding}}      \\ \hdashline
CLIP                    & 29.2          & 54.6          & 29.9          & \multicolumn{1}{c|}{37.9}          & 66.6          & 93.5          & 87.9          & 82.7          \\
KeepFITV1                                             & 37.6          & 70.6          & \textbf{35.0} & \multicolumn{1}{c|}{47.7}          & 63.9          & \textbf{97.9} & \textbf{94.4} & 85.4 \\
\rowcolor[HTML]{DCDCDC}KeepFITV2               & \textbf{38.5} & \textbf{72.0} & 33.8          & \multicolumn{1}{c|}{\textbf{48.1}} & \textbf{70.7} & 97.3          & 93.0          & \textbf{87.0} \\ \hline
\end{tabular}
\end{table}

\begin{table*}[]
\caption{Few-Shot Classification Performance in OCT Modality across Two Datasets. (\%)}
\label{OCT_fewshot}
\setlength{\tabcolsep}{0.5pt}
\renewcommand\arraystretch{1.2}
\begin{tabular}{ccccccccccccccccccccccccccccc}
\hline
\multirow{3}{*}{Model} & \multicolumn{9}{c|}{Clipadapter}   & \multicolumn{9}{c|}{Tipadapter}     & \multicolumn{9}{c|}{Tipadapter-f}      & \multirow{3}{*}{AVG} \\
   & \multicolumn{3}{c}{1}                         & \multicolumn{3}{c}{5}                         & \multicolumn{3}{c|}{10}                                            & \multicolumn{3}{c}{1}                         & \multicolumn{3}{c}{5}                         & \multicolumn{3}{c|}{10}                                            & \multicolumn{3}{c}{1}                         & \multicolumn{3}{c}{5}                         & \multicolumn{3}{c|}{10}                                            &                      \\ \cline{2-28}
                       & {\fontsize{7pt}{12pt}\selectfont ACC}   & {\fontsize{7pt}{12pt}\selectfont AUC}   & {\fontsize{7pt}{12pt}\selectfont AUPR}  & {\fontsize{7pt}{12pt}\selectfont ACC}   & {\fontsize{7pt}{12pt}\selectfont AUC}   & {\fontsize{7pt}{12pt}\selectfont AUPR}  & {\fontsize{7pt}{12pt}\selectfont ACC}   & {\fontsize{7pt}{12pt}\selectfont AUC}   & \multicolumn{1}{c|}{{\fontsize{7pt}{12pt}\selectfont AUPR}}  & {\fontsize{7pt}{12pt}\selectfont ACC}   & {\fontsize{7pt}{12pt}\selectfont AUC}   & {\fontsize{7pt}{12pt}\selectfont AUPR}  & {\fontsize{7pt}{12pt}\selectfont ACC}   & {\fontsize{7pt}{12pt}\selectfont AUC}   & {\fontsize{7pt}{12pt}\selectfont AUPR}  & {\fontsize{7pt}{12pt}\selectfont ACC}   & {\fontsize{7pt}{12pt}\selectfont AUC}   & \multicolumn{1}{c|}{{\fontsize{7pt}{12pt}\selectfont AUPR}}  & {\fontsize{7pt}{12pt}\selectfont ACC}   & {\fontsize{7pt}{12pt}\selectfont AUC}   & {\fontsize{7pt}{12pt}\selectfont AUPR}  & {\fontsize{7pt}{12pt}\selectfont ACC}   & {\fontsize{7pt}{12pt}\selectfont AUC}   & {\fontsize{7pt}{12pt}\selectfont AUPR}  & {\fontsize{7pt}{12pt}\selectfont ACC}   & {\fontsize{7pt}{12pt}\selectfont AUC}   & \multicolumn{1}{c|}{{\fontsize{7pt}{12pt}\selectfont AUPR}}  &                      \\ \hline
\multicolumn{29}{c}{OCTDL} \\ \hline
\multicolumn{29}{l}{\textit{Generalist VLPs for biomedical understanding}}  \\ \hdashline
BiomedCLIP             & 27.4          & 65.8          & 25.1          & 39.2          & 75.0          & 33.4          & 48.6          & 82.6          & \multicolumn{1}{c|}{43.0}          & 18.5          & 56.4          & 18.8          & 19.1          & 58.3          & 19.5          & 19.7          & 60.7          & \multicolumn{1}{c|}{20.5}          & 17.5          & 55.9          & 19.1          & 19.5          & 60.6          & 20.3          & 21.7          & 63.7          & \multicolumn{1}{c|}{23.2}          & 38.3                 \\
PubMedCLIP             & 23.1          & 60.8          & 22.5          & 34.4          & 70.5          & 31.2          & 43.7          & 77.2          & \multicolumn{1}{c|}{38.8}          & 15.9          & 54.1          & 17.5          & 16.0          & 58.1          & 18.8          & 16.2          & 62.4          & \multicolumn{1}{c|}{21.0}          & 15.1          & 53.6          & 18.2          & 17.0          & 61.4          & 21.2          & 24.7          & 69.6          & \multicolumn{1}{c|}{25.7}          & 36.6                 \\ \hline
\multicolumn{29}{l}{Specialist VLPs for retinal understanding} \\ \hdashline
CLIP                    & 40.4          & 79.8          & 40.4          & 55.4          & 88.1          & 54.1          & 58.4          & 90.9          & \multicolumn{1}{c|}{58.9}          & 29.6          & 55.0          & 27.5          & 30.3          & 57.3          & 28.9          & 31.2          & 59.5          & \multicolumn{1}{c|}{30.5}          & 30.0          & 56.1          & 27.6          & 32.3          & 63.4          & 33.4          & 41.9          & 68.7          & \multicolumn{1}{c|}{40.8}          & 48.5                 \\
KeepFITV1                                             & 41.3          & 79.7          & 40.5          & 52.3          & 86.8          & 54.1          & 55.7          & 90.5          & \multicolumn{1}{c|}{59.6}          & 37.8          & 68.8          & \textbf{34.3} & 38.4          & 70.5          & 25.1          & \textbf{39.2} & 72.1          & \multicolumn{1}{c|}{\textbf{36.3}} & \textbf{38.5} & 69.3          & \textbf{34.6} & 38.7          & 72.6          & 36.2          & 41.1          & 78.2          & \multicolumn{1}{c|}{40.1}          & 53.0                 \\
\rowcolor[HTML]{DCDCDC}KeepFITV2              & \textbf{46.2} & \textbf{81.3} & \textbf{46.6} & \textbf{57.3} & \textbf{89.4} & \textbf{59.8} & \textbf{61.8} & \textbf{92.3} & \multicolumn{1}{c|}{\textbf{66.8}} & \textbf{37.8} & \textbf{70.0} & 33.5          & \textbf{38.5} & \textbf{71.7} & \textbf{34.2} & 38.8 & \textbf{73.5} & \multicolumn{1}{c|}{35.0}          & 37.6          & \textbf{70.3} & 33.3          & \textbf{41.6} & \textbf{77.0} & \textbf{37.6} & \textbf{43.2} & \textbf{82.6} & \multicolumn{1}{c|}{\textbf{43.0}} & \textbf{55.6}       \\ \hline
\multicolumn{29}{c}{OCTID}  \\ \hline
\multicolumn{29}{l}{\textit{Generalist VLPs for biomedical understanding}}    \\ \hdashline
BiomedCLIP             & 54.1          & 81.1          & 58.0          & 62.2          & 89.2          & 68.5          & 68.6          & 91.8          & \multicolumn{1}{c|}{73.5}          & 18.6          & 59.0          & 32.9          & 19.9          & 59.6          & 33.7          & 22.9          & 61.1          & \multicolumn{1}{c|}{35.7}          & 18.4          & 61.0          & 34.9          & 23.2          & 60.5          & 36.0          & 26.5          & 64.9          & \multicolumn{1}{c|}{39.0}          & 50.2                 \\
PubMedCLIP             & 31.0          & 69.5          & 41.8          & 58.8          & 86.7          & 64.9          & 64.9          & 89.8          & \multicolumn{1}{c|}{71.3}          & 21.5          & 53.0          & 24.8          & 26.9          & 61.6          & 29.3          & 31.6          & 71.1          & \multicolumn{1}{c|}{37.1}          & 19.2          & 55.4          & 26.5          & 33.3          & 68.8          & 36.8          & 43.8          & 79.1          & \multicolumn{1}{c|}{51.7}          & 50.0                 \\ \hline
\multicolumn{29}{l}{\textit{Specialist VLPs for retinal understanding}}   \\ \hdashline
CLIP                    & \textbf{91.8} & \textbf{97.8} & \textbf{95.5} & 92.4          & \textbf{99.3} & \textbf{96.4} & 92.5          & 98.4          & \multicolumn{1}{c|}{\textbf{96.7}} & 58.2          & 93.6          & 77.9          & 65.0          & 92.4          & 81.6          & 72.9          & 94.3          & \multicolumn{1}{c|}{84.6}          & 59.8          & 91.6          & 77.7          & 70.3          & 95.7          & 86.3          & 80.0          & 97.4          & \multicolumn{1}{c|}{90.6}          & 86.3                 \\
KeepFITV1                                             & 91.0          & 97.9          & 94.7          & \textbf{92.8} & 98.4          & 95.6          & \textbf{93.6} & 97.9          & \multicolumn{1}{c|}{95.3}          & 67.3          & \textbf{97.3} & \textbf{93.5} & 73.9          & 97.6 & 94.1 & 80.9          & 97.9          & \multicolumn{1}{c|}{94.6}          & 68.3          & \textbf{98.0} & \textbf{95.0} & 79.2          & 97.5          & 94.3          & 88.9          & 98.4          & \multicolumn{1}{c|}{95.5}          & 91.5
                 \\
\rowcolor[HTML]{DCDCDC}KeepFITV2              & 89.6          & \textbf{97.8} & 94.6          & 92.4          & 98.8          & \textbf{96.4} & 92.8          & \textbf{98.5} & \multicolumn{1}{c|}{96.6}          & \textbf{76.4} & 97.0          & 92.7          & \textbf{82.4} & \textbf{97.9} & \textbf{94.2}          & \textbf{85.2} & \textbf{98.7} & \multicolumn{1}{c|}{\textbf{95.4}} & \textbf{76.5} & 97.6          & 93.5          & \textbf{86.5} & \textbf{98.6} & \textbf{95.8} & \textbf{90.2} & \textbf{98.9} & \multicolumn{1}{c|}{\textbf{96.6}} & \textbf{93.0}                 \\ \hline
\end{tabular}
\end{table*}

Finally, we examine the generalization capability and transferability of different vision-language pretraining models under zero-shot, few-shot, and linear probing settings in OCT modality. Table~\ref{OCT_zeroshot}, Table~\ref{OCT_fewshot}, and Fig.~\ref{FFA_OCT_lp} demonstrate the comparison results. Similarly, KeepFIT V2 achieves the highest average score on OCTDL and OCTID in three scenarios. The enhancement can be attributed to the combination of high-level semantics-oriented and low-level appearance-oriented knowledge injection, which facilitates the vision-language alignment and feature understanding. 

Experiments across three different modalities illustrate that the high-quality expert knowledge contained within the MM-Retinal V2 dataset significantly benefits the training of foundation models in fundus image analysis. In addition, these experiments also validate the effectiveness of the proposed KeepFIT V2, which successfully uses only a minimal amount of elite image-text data as a spark to achieve comparable performance to those vision-language pretraining models trained on large-scale private image-text pairs.

\subsection{Ablation Study}
\label{ablation_study}

\begin{table*}[]
\caption{Ablation Study in CFP Modality across Zero-Shot, Few-Shot, and Linear Probing. For each dataset, the average values of all metrics under each setting are presented. KI refers to knowledge injection (\%)}
\label{ablation_study_table}
\setlength{\tabcolsep}{3.4pt}
\renewcommand\arraystretch{1.2}
\begin{tabular}{ccc|ccccc|ccccc|ccccc}
\hline
\multirow{2}{*}{\begin{tabular}[c]{@{}c@{}}Semantic \\ KI \end{tabular}} & \multirow{2}{*}{\begin{tabular}[c]{@{}c@{}}Textual\\ Pretraining\end{tabular}} & \multirow{2}{*}{\begin{tabular}[c]{@{}c@{}}Appearance \\ KI\end{tabular}} & \multicolumn{5}{c|}{Zero-Shot}        & \multicolumn{5}{c|}{Few-Shot}         & \multicolumn{5}{c}{Linear Probing}   \\
&     &     & {\fontsize{7pt}{12pt}\selectfont REFUGE} & {\fontsize{7pt}{12pt}\selectfont ODIR} & {\fontsize{7pt}{12pt}\selectfont Retina}  & {\fontsize{7pt}{12pt}\selectfont 
AMD}  & {\fontsize{7pt}{12pt}\selectfont AVG}  & {\fontsize{7pt}{12pt}\selectfont REFUGE} & {\fontsize{7pt}{12pt}\selectfont ODIR}  & {\fontsize{7pt}{12pt}\selectfont Retina} & {\fontsize{7pt}{12pt}\selectfont AMD}  & {\fontsize{7pt}{12pt}\selectfont AVG}  & {\fontsize{7pt}{12pt}\selectfont REFUGE} & {\fontsize{7pt}{12pt}\selectfont ODIR}  & {\fontsize{7pt}{12pt}\selectfont APTOS} & {\fontsize{7pt}{12pt}\selectfont FIVES} & {\fontsize{7pt}{12pt}\selectfont AVG}                         \\ \hline
{\scalebox{0.7}{\XSolidBrush}} & {\scalebox{0.7}{\XSolidBrush}}    & {\scalebox{0.7}{\XSolidBrush}}    & 89.3   & 68.2 & 49.9   & 74.9 & 70.6 & 87.4   & 73.8 & 51.9   & 77.1 & 72.6 & 88.3   & 94.3 & 83.0  & 91.9  & 89.4                      \\
{\scalebox{1.1}{\checkmark}}    & {\scalebox{0.7}{\XSolidBrush}}  & {\scalebox{0.7}{\XSolidBrush}}   & 89.4   & 87.2 & 57.4   & 83.8 & 79.5 & 87.8   & 90.6 & 63.3   & 84.6 & 81.6 & 90.5   & 95.7 & 82.4  & 92.5  & 90.3                   \\
{\scalebox{1.1}{\checkmark}}    & {\scalebox{1.1}{\checkmark}}    & {\scalebox{0.7}{\XSolidBrush}}     & 89.7   & \textbf{91.3} & 58.7   & 80.3 & 80.0 & 88.7   & \textbf{92.2} & \textbf{66.2}   & 80.3 & 81.9 & 91.0   & \textbf{96.3} & 83.0  & 92.4  & 90.7                   \\
\rowcolor[HTML]{DCDCDC}(Ours) {\scalebox{1.1}{\checkmark}}    & {\scalebox{1.1}{\checkmark}}      & {\scalebox{1.1}{\checkmark}}       & \textbf{92.8}   & 87.2 & \textbf{61.1}   & \textbf{85.9} & \textbf{81.8} & \textbf{90.1}   & 90.8 & 64.4   & \textbf{86.0} & \textbf{82.8} & \textbf{91.5}   & 95.6 & \textbf{88.3}  & \textbf{95.3}  & \textbf{92.7}                    \\ \hline
\end{tabular}
\end{table*}

In this section, we validate the effectiveness of each module in KeepFIT V2. Table~\ref{ablation_study_table} presents the average score of each downstream dataset under zero-shot, few-shot, and linear probing settings in CFP modality. 
The Semantic KI and Appearance KI represent semantics-oriented and appearance-oriented expert knowledge extraction, respectively, along with their associated expert knowledge refinements. These elements collectively constitute the hybrid image-text knowledge injection. Textual Pretraining refers to the preliminary textual knowledge pretraining in Section ~\ref{textual_knowledge_pretraining}. 
A primary observation from the results is that the removal of any module leads to a performance decline to varying degrees, highlighting that all the modules contribute to the performance improvement.
Notably, the results of the second and the last rows demonstrate the crucial role of the hybrid image-text knowledge injection module in achieving strong knowledge transfer from MM-Retinal V2 to categorical public datasets. 
Although a slight decline is observed in minor cases, overall averaged performance improves across all datasets and settings.

%%%%%%%%%%%%%%%%%%%%%%%%%%%%%%% Conclusion %%%%%%%%%%%%%%%%%%%%%%%%%%%%%%%%%%%%%%%
\section{Conclusion}
In this work, we construct MM-Retinal V2, a high-quality image-text dataset encompassing CFP, FFA, and OCT modalities, and covering over 96 fundus diseases and abnormalities. 
Enabled by MM-Retinal V2 and public categorically-labeled datasets, we propose KeepFIT V2, a vision-language foundation model for retinal image analysis. 
KeepFIT V2 effectively incorporates expert knowledge from MM-Retinal V2 into foundation model pretraining through preliminary textual pretraining and hybrid image-text knowledge injection, which leverages a combination of high-level semantic features from contrastive learning and low-level appearance features from generative learning to enhance its performance.
Moreover, KeepFIT V2 provides a novel approach to building retinal foundation model with the elite MM-Retinal V2 spark instead of relying on large-scale private image-text data, while still delivering competitive performance. 
Our proposed knowledge spark spreading pretraining scheme is not only effective for retinal foundation model pretraining but can also be broadly applied to other medical foundation models encountering the same challenge of limited image-text data. This highlights the versatility and generalizability of our scheme, providing a promising solution for advancing vision-language pretraining across diverse medical imaging domains.

%%%%%%%%%%%%%%%%%%%%%%%%%%%%%%% Acknowledment %%%%%%%%%%%%%%%%%%%%%%%%%%%%%%%%%%%%%%%
\section*{Acknowledgments}
This work was supported by the National Natural Science Foundation of China (Nos. 62476054, and 62172228).

%%%%%%%%%%%%%%%%%%%%%%%%%%%%%%% References %%%%%%%%%%%%%%%%%%%%%%%%%%%%%%%%%%%%%%%
% \section*{References}
\bibliographystyle{splncs04}
\bibliography{BibTex.bib}

\end{document}